\documentclass[10pt,twocolumn,letterpaper]{article}

\usepackage[pagenumbers]{cvpr} %

\usepackage{graphicx}
\usepackage{amsmath}
\usepackage{amssymb}
\usepackage{booktabs}
\usepackage{paralist}

\usepackage{xr-hyper}
\usepackage[pagebackref,breaklinks,colorlinks,citecolor=citecolor]{hyperref}
\usepackage{xspace}

\usepackage[capitalize]{cleveref}

\usepackage{xcolor}

\usepackage{soul}
\usepackage{subcaption}

\usepackage{lipsum}  
\usepackage{pifont}
\usepackage{enumitem}
\usepackage{physics}

\crefname{section}{Sec.}{Secs.}
\Crefname{section}{Section}{Sections}
\Crefname{table}{Table}{Tables}
\crefname{table}{Tab.}{Tabs.}

\setlist{nosep}

\definecolor{citecolor}{RGB}{30,102,235}

\DeclareMathOperator{\spn}{span}
\DeclareMathOperator{\prj}{proj}

\makeatletter
\DeclareRobustCommand\onedot{\futurelet\@let@token\@onedot}
\def\@onedot{\ifx\@let@token.\else.\null\fi\xspace}

\def\eg{\emph{e.g}\onedot} 
\def\ie{\emph{i.e}\onedot}

\def\etal{et~al\onedot}
\makeatother

\definecolor{mydarkblue}{rgb}{0,0.08,1}
\definecolor{mydarkgreen}{rgb}{0.02,0.6,0.02}
\definecolor{mydarkorange}{rgb}{0.40,0.2,0.02}
 \definecolor{mycyan}{rgb}{0.02,0.6,0.42}

\newif\ifdraft
\draftfalse

\ifdraft
    \newcommand{\ync}[1]{{\color{blue}\textbf{YN:} #1}}
    \newcommand{\dcc}[1]{{\color{red}\textbf{Danny:} #1}}
    \newcommand{\tae}[1]{{\color{orange}\textbf{Tae:} #1}}
    \newcommand{\mgc}[1]{{\color{mydarkgreen}\textbf{Michael:}#1}}
    \newcommand{\esc}[1]{{\color{mycyan}\textbf{ES:} #1}}
    \newcommand{\MG}[1]{\textcolor{teal}{MG: #1}}
    \newcommand{\jzc}[1]{\color{yellow}{JZ: #1}}

    \newcommand{\yn}[1]{{\color{blue}#1}}

\else
    \newcommand{\ync}[1]{}
    \newcommand{\dcc}[1]{}
    \newcommand{\mgc}[1]{}
    \newcommand{\MG}[1]{}
    \newcommand{\tae}[1]{}
    \newcommand{\jzc}[1]{}
    \newcommand{\esc}[1]{}
    
    \newcommand{\yn}[1]{{\color{black}#1}}

\fi

\newcommand{\myparagraph}[1]{\vspace{-7pt}\paragraph{#1}}

\newcommand{\notation}[1]{\ensuremath{#1}\xspace}

\newcommand{\w}{\notation{\mathcal{Z}}}
\newcommand{\di}{\notation{\mathcal{D}_i}}
\newcommand{\li}{\notation{\mathcal{L}_i}}
\newcommand{\dsrc}{\notation{\mathcal{D}_{\text{src}}}}
\newcommand{\gsrc}{\notation{G_{\text{src}}}}
\newcommand{\lsrc}{\notation{\mathcal{L}_{\text{src}}}}

\newcommand{\wi}{\notation{\mathcal{Z}_i}}
\newcommand{\wj}{\notation{\mathcal{Z}_j}}
\newcommand{\wbase}{\notation{\mathcal{Z}_{\text{base}}}}
\newcommand{\gt}{\notation{G^{+}}}
\newcommand{\gi}{\notation{G_i}}

\newlist{todolist}{itemize}{2}
\setlist[todolist]{label=$\square$}

\def\cvprPaperID{903} %
\def\confName{CVPR}
\def\confYear{2023}

\makeatletter
    \newcommand*{\addFileDependency}[1]{%
      \typeout{(#1)}
      \@addtofilelist{#1}
      \IfFileExists{#1}{}{\typeout{No file #1.}}
    }
\makeatother

\begin{document}

\title{Domain Expansion of Image Generators}

\author{
 \hspace{-6mm}Yotam Nitzan$^{1,2}$  \and
 Michaël Gharbi$^{1}$  \and
 Richard Zhang$^1$  \and
 Taesung Park$^{1}$ \and 
 Jun-Yan Zhu$^{3}$  \and
 Daniel Cohen-Or$^2$  \and
 Eli Shechtman$^1$ \and
 \\
 $^{1}$ Adobe Research \qquad $^2$ Tel-Aviv University \qquad $^3$ Carnegie Mellon University 
}

\twocolumn[{%
    \maketitle
    \begin{center}
        \centering
        \includegraphics[width=0.95\linewidth]{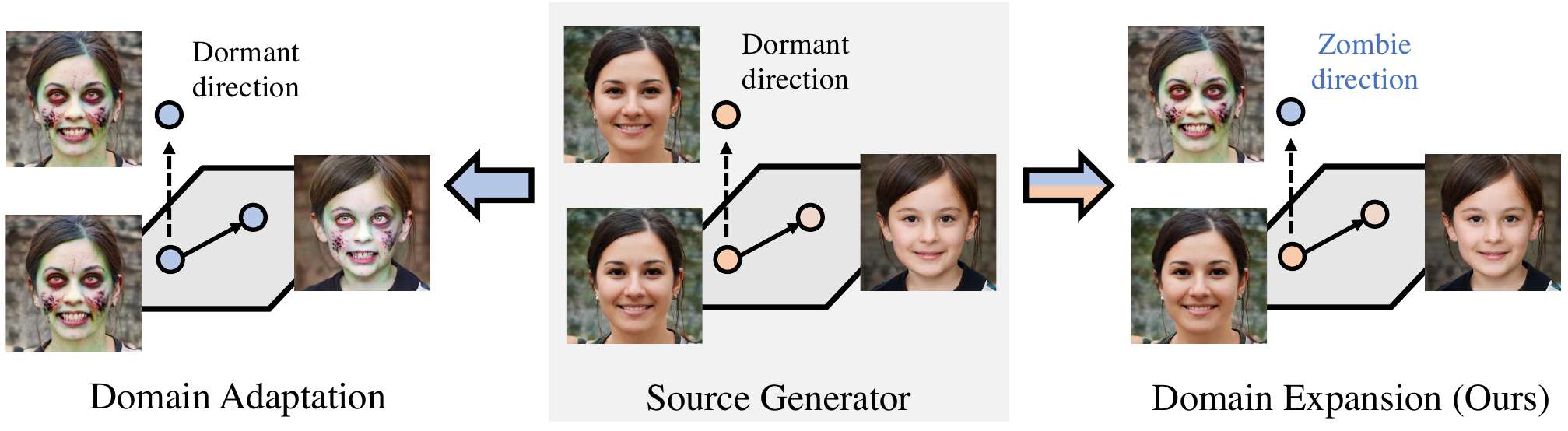} 
        \captionof{figure}{
        (center) Traversing the latent space of generative models along some directions changes the image significantly while traversing others has no perceptible effect. We call directions of the latter type dormant. (left) Domain adaptation methods, transform the entire generator from a source domain to a target domain, indicated by the color blue. (right) We introduce an approach for a new task -- domain expansion. Instead of fully transforming the generator, we expand it to include new data domains. Our method learns to represent the new domain in a disentangled manner by repurposing a single dormant direction.
        }
        \label{fig:latent_space_structure}
    \end{center}
}]

\begin{abstract}
\vspace{-3.5mm}
Can one inject new concepts into an already trained generative model, while respecting its existing structure and knowledge?
We propose a new task -- domain expansion -- to address this.
Given a pretrained generator and novel (but related) domains, we expand the generator to jointly model all domains, old and new, harmoniously.
First, we note the generator contains a meaningful, pretrained latent space. Is it possible to minimally perturb this hard-earned representation, while maximally representing the new domains? Interestingly, we find that the latent space offers unused, ``dormant'' directions, which do not affect the output. This provides an opportunity: By ``repurposing'' these directions, we can represent new domains without perturbing the original representation.
In fact, we find that pretrained generators have the capacity to add several -- even hundreds -- of new domains!
Using our expansion method, one ``expanded'' model can supersede numerous domain-specific models, without expanding the model size. 
Additionally, a single expanded generator natively supports smooth transitions between domains, as well as composition of domains.

Code and project page available \href{https://yotamnitzan.github.io/domain-expansion/}{here}.

\end{abstract}

\section{Introduction}

Recent \emph{domain adaptation} techniques piggyback on the tremendous success of modern generative image models~\cite{karras2019style,brock2018large,vahdat2021score,rombach2022high}, by adapting a pretrained generator so it can generate images from a new target domain. 
Oftentimes, the target domain is defined with respect to the source domain~\cite{gal2021stylegan,ojha2021few,nitzan2022mystyle}, e.g., changing the ``stylization'' from a photorealistic image to a sketch.
When such a relationship holds, domain adaptation typically seeks to preserve the factors of variations learned in the source domain, and transfer them to the new one (e.g., making the human depicted in a sketch smile based on the prior from a face generator).
With existing techniques, however, the adapted model loses the ability to generate images from the original domain.

In this work, we introduce a novel task --- \emph{domain expansion}.
Unlike domain adaptation, we aim to \textit{augment} the space of images a single model can generate, without overriding its original behavior (see \cref{fig:latent_space_structure}).
Rather than viewing similar image domains as disjoint data distributions, we treat them as different modes in a joint distribution.
As a result, the domains share a semantic prior inherited from the original data domain.
For example, the inherent factors of variation for photorealistic faces, such as pose and face shape, can equally apply to the domain of ``zombies''.

To this end, we carefully structure the model training process for expansion, respecting the original data domain. It is well-known that modern generative models with low-dimensional latent spaces offer an intriguing, emergent property -- through training, the latent spaces represent the factors of variation, in a linear and interpretable manner~\cite{radford2015unsupervised,brock2018large,jahanian2019steerability,karras2019style,harkonen2020ganspace,vahdat2020nvae,preechakul2022diffusion,vahdat2021score}.
We wish to extend this advantageous behavior and represent the new domains along linear and disentangled directions. Interestingly, it was previously shown that many latent directions have insignificant perceptible effect on generated images~\cite{harkonen2020ganspace}. Taking advantage of this finding, we repurpose such directions to represent the new domains.

In practice, we start from an orthogonal decomposition of the latent space~\cite{shen2020closedform} and identify a set of low-magnitude directions that have no perceptible effect on the generated images, which we call \emph{dormant}.
To add a new domain, we select a dormant direction to repurpose. Its orthogonal subspace, which we call \textit{base subspace}, is sufficient to represent the original domain~\cite{harkonen2020ganspace}. 
We aim to repurpose the dormant direction such that traversing it would now cause a transition between the original and the new domain. Specifically, the transition should be disentangled from the original domain's factors of variation.
To this end, we define a \emph{repurposed} affine subspace by transporting the base subspace along the chosen dormant direction, as shown in~\cref{fig:training}. 
We capture the new domain by applying a domain adaptation method, transformed to operate only on latent codes sampled from the repurposed subspace.
A regularization loss is applied on the base subspace to ensure that the original domain is preserved. 
The original domain's factors of variation are implicitly preserved due to the subspaces being parallel and the latent space being disentangled.
For multiple new domains, we simply repeat this procedure across multiple dormant directions.

\begin{figure}
    \centering
    \includegraphics[width=\linewidth]{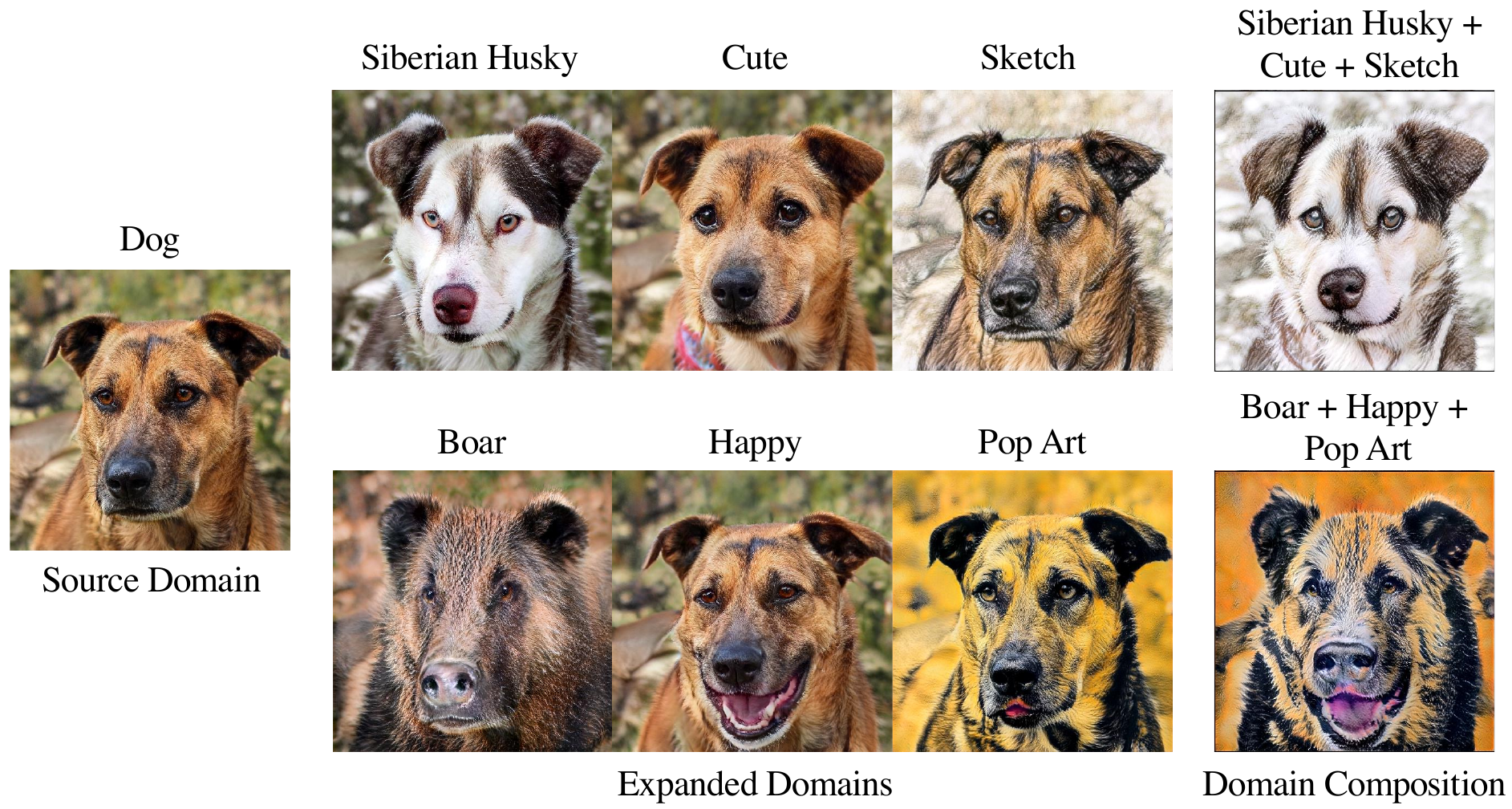}
    \caption{Example of a \textit{domain expansion} result. Starting from dogs as the source domain, we expand a \emph{single} generator to model new domains such as facial expressions, breeds of dogs and other animals, and artistic styles. Finally, as the representations are disentangled, the expanded generator is able to generalize and compose the different domains, although they were never seen jointly in training.}
    \vspace{-3mm}
    \label{fig:composition_teaser}
\end{figure}

We apply our method to two generators architectures, StyleGAN \cite{karras2020analyzing} and Diffusion Autoencoder~\cite{preechakul2022diffusion}, trained over several datasets, and expand the generator with \textit{hundreds} of new factors of variation.
Crucially, we show our expanded model simultaneously generates high-quality images from both original and new domains, comparable to specialized, domain-specific generators.
Thus, a single expanded generator supersedes hundreds of adapted generators, facilitating the deployment of generative models for real-world applications. 
We additionally demonstrate that the new domains are learned as global and disentangled factors of variation, alongside existing ones.
This enables fine-grained control over the generative process and paves the way to new applications and capabilities, \eg, compositing multiple domains (See \cref{fig:composition_teaser}).
Finally, we conduct a detailed analysis of key aspects of our method, such as the effect of the number of newly introduced domains, thus shedding light on our method and, in the process, on the nature of the latent space of generative models.

To summarize, our contributions are as follows:
\begin{itemize}[topsep=0pt,leftmargin=*]
    \item We introduce a new task -- domain expansion of a pretrained generative model.
    \item We propose a novel latent space structure that is amenable to representing new knowledge in a disentangled manner, while maintaining existing knowledge intact.
    \item We present a simple paradigm transforming domain adaptation methods into domain expansion methods.
\item We demonstrate successful domain expansion to hundreds of new domains and illustrate its advantage over domain adaptation methods.
\end{itemize}

\section{Related Work}

\myparagraph{Fine-tuning generative models.}

Starting from a generator pretrained on a source domain and training it for a target domain, often called fine-tuning, is a common technique applied for various purposes and settings.

Some works wish to model only the target domain. In which case, the pretrained model is leveraged simply as an efficient initialization, shortening the training time, and improving image quality~\cite{Karras2020ada,zhao2020differentiable,yang2021data,mo2020freeze,kumari2022ensembling}.
Others, wish to learn the target domain alongside the source domain, in a setting called continuous learning, and propose methods to ensure that the source domain is not forgotten~\cite{seff2017continual,wu2018memory}. Although in a single generator, the domains are modeled separately,  each as its own class.

A prominent line of works have sought to make the target domain inherit knowledge from the source domain~\cite{gal2021stylegan,ojha2021few,zhu2021mind,kim2022diffusionclip,bau2020rewriting,wang2022rewriting,li2020few,wang2021sketch,roich2021pivotal,bau2020semantic,pan2021exploiting,nitzan2022mystyle,ruiz2022dreambooth}. This approach allows generalization beyond the target domain per-se and is especially useful when training data is scarce.

Our work similarly involves fine-tuning, but for a novel purpose. Our perspective is that, since the target domain is introduced with knowledge from the source domain -- it is in essence, an expansion of it. Therefore, in contrast to the aforementioned works, we aim to model the domains jointly. The proposed method does not replace previous fine-tuning methods, but allows applying them jointly.

\myparagraph{Latent directions in generative models.}

Generative models learn to represent the factors of variation of observed data in their latent space.
Disentangled representations are especially useful as they facilitate intuitive control over the generative process.
With recent architectures, disentanglement miraculously emerges without intervention~\cite{karras2019style,preechakul2022diffusion,park2020swapping,vahdat2020nvae}.
In such models, disentanglement is manifested through the existence of linear latent directions, each ideally controlling a single factor of variation.

Due to the spontaneous emergence of such directions, many works have been proposed to identify them after the model has been trained~\cite{shen2020closedform,shen2020interfacegan,wu2020stylespace,patashnik2021styleclip,harkonen2020ganspace,voynov2020unsupervised,spingarn2020gan} and used them for downstream applications, most commonly semantic image editing. At the same time, it has also been observed that some latent directions have no perceptible effect on the generated images~\cite{harkonen2020ganspace,wu2021stylealign}. These directions, which we call \emph{dormant}, were not previously leveraged for any purpose.

In this work, we rely on existing methods to factorize the latent space into such linear directions. As we aim to expand the pretrained generator to additional domains, we decide to explicitly encode the ``new knowledge'' along the dormant directions, while keeping other directions intact. This design ensures that the original domain is preserved and that the different domains are represented in a disentangled fashion.

\section{Method}
\label{sec:method}

We start with a pretrained generator \gsrc that maps from latent codes $z \in \mathcal{Z} \subseteq \mathbb{R}^D$ to images in a source domain \dsrc,
and a set of $N$ domain adaptation tasks, each defined by a loss function \li, $i\in\{1,\ldots,N\}$.
In domain adaptation, fine-tuning \gsrc to minimize \li yields a generator \gi that generates images from the new domain \di.
In contrast, our goal is domain expansion, which aims at training a single expanded generator \gt that can \emph{simultaneously} model all the new domains $\cup_{i=1}^N\mathcal{D}_i$, along with the original domain \dsrc.
We want to ensure that the new domains \di are disentangled from each other and also share the factors of variation from the source domain, which remain intact.

\begin{figure}
    \centering
    \includegraphics[width=\linewidth]{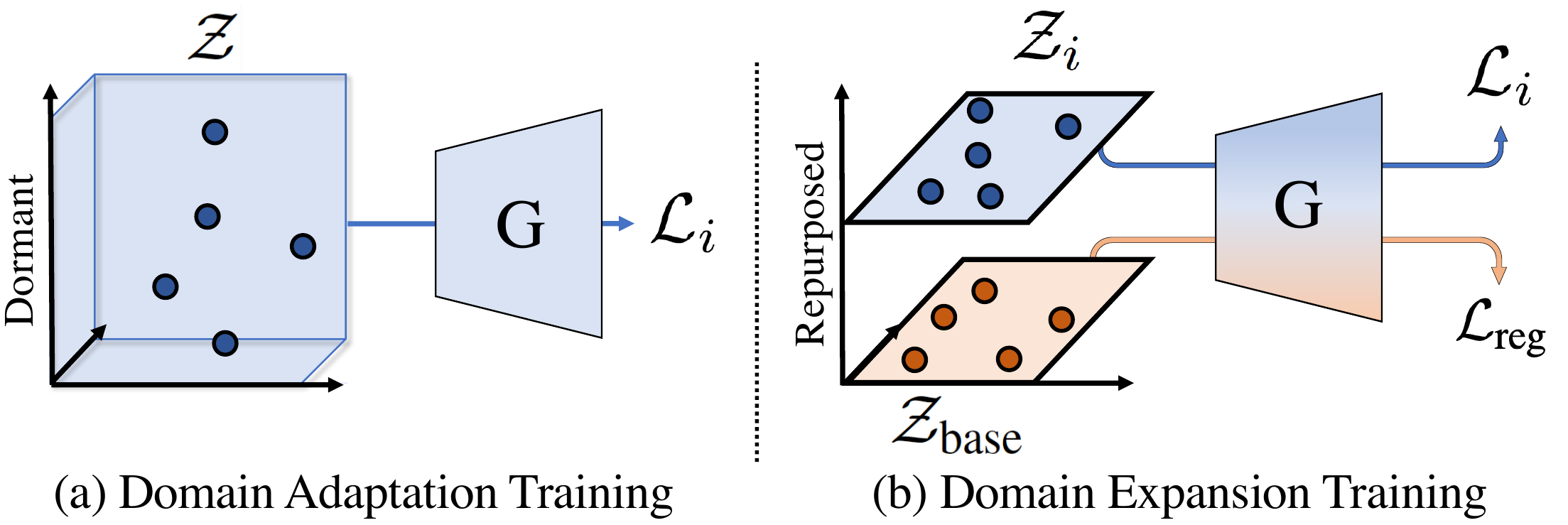}
    \caption{
    Our method transforms a domain adaptation task to a domain expansion task.
    (a) Generator $G$ is optimized to satisfy the loss $\mathcal{L}_i$ for every latent code in space. The entire generator and latent space now represent the new domain, indicated with the color blue. %
    (b) Generator $G$ is optimized to satisfy the same loss, $\mathcal{L}_i$, only on a subspace \wi, dedicated to the new domain. Simultaneously, $G$ is optimized to satisfy a regularization term $\mathcal{L}_\text{reg}$ on a parallel subspace, \wbase, ensuring the original knowledge is preserved there. The generator and latent space now represent both domains, indicated by being colored both blue and orange. The latent direction between the two spaces was originally dormant in generator $G$, and now represents a transition between the domains.
    }
    \vspace{-4mm}
    \label{fig:training}
\end{figure}

Our solution is to partition the latent space into disjoint subspaces, one for each new domain, and to restrict the effect of each domain adaptation to the corresponding subspace.
To this end, we endow the latent space with an explicit structure that supports domain expansion (\cref{subsec:latent_structure}), and
optimize each domain adaptation loss only using latents from specific subspace reserved for the new domain (\cref{subsec:adjusting}).
Our decomposition reserves a base subspace for the original domain \dsrc, on which we impose a regularization objective to maintain the behavior of the source generator (\cref{subsec:regularization}).
\cref{fig:training} gives an overview of our domain expansion algorithm.

\subsection{Structuring the Latent Space for Expansion}
\label{subsec:latent_structure}

Modern generative models conveniently learn to represent the factors of variation along linear latent directions, in a completely unsupervised manner~\cite{karras2019style,peebles2020hessian,vahdat2020nvae,preechakul2022diffusion}. We decide to explicitly extend this model by structuring the latent space such that the effect defined by an adaptation task would be represented along a single linear direction. Formally, there should exist some scalar $s$ and latent direction $v_i$, for which images generated from $\gt(z), \gt(z + sv_i)$, relate to each other as the corresponding images from the source and adapted generators $G_\text{src}(z)$, $G_i(z)$ do.

Concretely, following SeFA~\cite{shen2020closedform}, we obtain a semantic and orthogonal basis $V$ of the latent space from the right singular vectors (produced by SVD) of the very first generator layer, which acts on the latent space $\mathcal{Z}$.
With a similar factorization technique~\cite{harkonen2020ganspace}, it was observed that a relatively small subset of the basis vector is sufficient to represent most of the generators \gsrc's variability. Other basis vectors have barely any perceptible effect on the generated images. We find this to be the case with SeFA as well.
We refer to vectors with no perceptible effect as \emph{dormant}.

As the dormant directions do not affect the model's generation capabilities, they are available to be repurposed with new desired behavior. We thus choose to represent the domains \dsrc and \di in regions that are separated by only a dormant direction.

Formally, for each of the $N$ adaptation tasks, we dedicate a single dormant direction, $v_i$, that will be repurposed. The remaining directions $\{v_{N+1},\ldots,v_{D}\}$ will remain intact.
We finally define a subspace of $\mathcal{Z}$, dubbed the \textit{base subspace}, as
\begin{equation}
    \label{eq:base_space}
    \mathcal{Z}_{\text{base}} = \spn(v_{N+1},\ldots,v_D) + \overline{z}
\end{equation}
where $\overline{z}$ is the mean of the distribution over the latents used to train the generator.
Then, for each repurposed direction, $v_i$, we define a \textit{repurposed subspace} $ \mathcal{Z}_{i}$ that is the base subspace transported along direction $v_i$ by a predetermined scalar size $s$.
\begin{equation}
\wi = \wbase + s v_i.
\end{equation}
The choice of direction $v_i$ and scalar $s$ are discussed in~\cref{subsec:effect_of_dim,subsec:effect_of_s}.

Our domain expansion training procedure described hereafter will ensure subspace \wi is the only part of the latent space affected by the training objective \li, and is reserved to generate images from domain \di.
Intuitively, shifting the base subspace along direction $v_i$ aims to achieve two goals: 
\begin{inparaenum}
\item preserve the factors of variations inherited from \wbase, and
\item restrict the new factor of variation (corresponding to \di) to a single latent direction, $v_i$.
\end{inparaenum}

\subsection{From Domain Adaptation to Expansion}
\label{subsec:adjusting}

Having defined disjoint affine subspaces \wi of the latent space \w for our new domains \di,
we now describe how we constrain each domain adaptation objective \li to affect only the corresponding subspace.

The domain adaptation objective is applied to images generated from latent codes $z\in\w$, sampled from distribution $p(z)$ defined on the entire space \w.
Commonly the distribution is a Gaussian, or is derived from it~\cite{karras2019style} but some exceptions exist~\cite{nitzan2022mystyle,wang2022rewriting}.
Our strategy is to transform this sample distribution into one restricted to the affine subspace \wi.
We do so by projecting the samples from $p(z)$ onto \wi, using a standard orthogonal projection operator
\begin{equation}
\label{eq:repurposed_space}
\prj_{\mathcal{Z}_i}(z) = \sum_{j=N+1}^D (v_j^\top (z - \overline{z})) v_j + \overline{z} + s v_i.
\end{equation}

Denoting by $p_i$ the sampling distribution over \w for each of the new domains we seek to adapt,
the training loss over all tasks is defined as
\begin{equation}
\mathcal{L}_{\text{expand}} = \sum_{i=1}^N \, \mathbb{E}_{z \sim p_i (z)} \;
\mathcal{L}_i(G(\prj_{\mathcal{Z}_i}(z))).
\end{equation}

\subsection{Regularization}
\label{subsec:regularization}

\begin{figure}
    \centering
    \includegraphics[width=\linewidth]{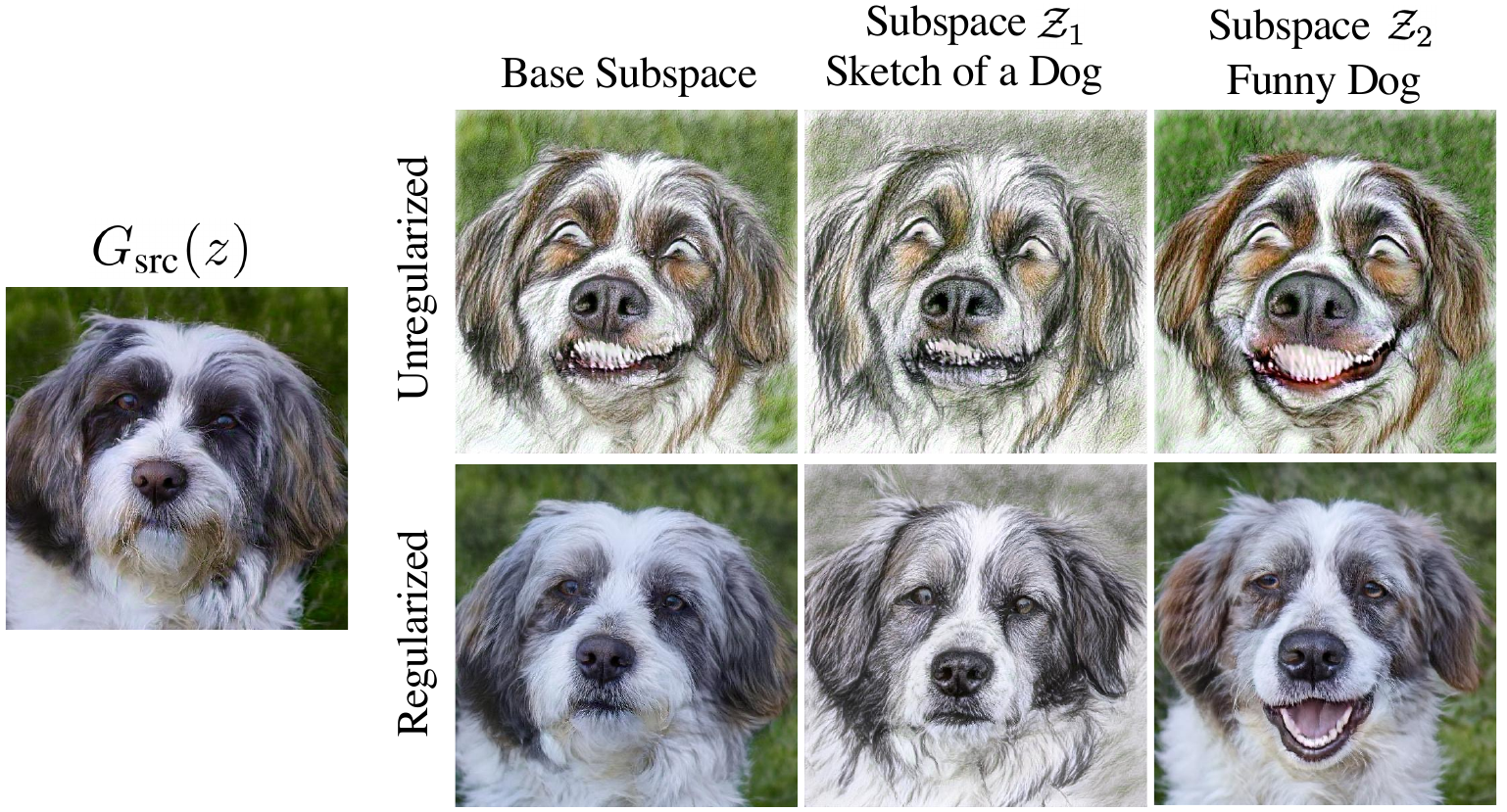}
    \caption{
    Regularization prevents leakage. Without regularization (top row), new factors of variation ``Sketch" and ``Funny" are leaking into the base subspace and the other repurposed subspace. 
    Note, for example, that the image from the base subspace is both a sketch and depicts a smiling dog.
    Our regularization, described in~\cref{eq:regularize_loss}, solves the issue (bottom row).
    }
    \vspace{-4mm}
    \label{fig:leakage}
\end{figure}

Optimizing $\mathcal{L}_{\text{expand}}$ lets us learn to generate data from the new domains \di within a single generator, but unfortunately it leaves the base subspace \wbase under-constrained and, therefore, does not guarantee it will remain unaltered during training.
In practice, we observe that the effect of $\mathcal{L}_i$ ``leaks'' outside \wi, causing \textit{catastrophic forgetting}~\cite{MCCLOSKEY1989109} in subspace \wbase, and undesirably affecting other subspaces \wj.
We show an example of this leakage in \cref{fig:leakage}.

To prevent this failure mode, we explicitly enforce the preservation of \gsrc's behavior over the base subspace \wbase by regularization.
We adopt two successful regularization techniques.
First, we keep optimizing the generator with the loss it was originally trained on, $\mathcal{L}_{\text{src}}$, which is known to mitigate forgetting~\cite{kirkpatrick2017overcoming}.
Second, we apply \textit{replay alignment}~\cite{wu2018memory}, which is a reconstruction loss 
that compares the output of a frozen copy of the source generator to that produced by our generator.
We use a weighted combination of an $L_2$ pixel loss and LPIPS~\cite{zhang2018unreasonable}:
\begin{equation}
\label{eq:recon_loss}
\begin{split}
    \mathcal{L}_{\text{recon}} = \lambda_{\text{lpips}} \mathcal{L}_{\text{lpips}}(G_{\text{src}}(z), G(z)) +
    \lambda_{L_2} \norm{G_{\text{src}}(z) - G(z)}_2,
\end{split}
\end{equation}
where $\lambda_{\text{lpips}} = \lambda_{L_2} = 10$ are weighting hyperparameters. 
Not only does replay alignment preserve the source domain \dsrc, it also has the added benefit of aligning \gt to the source generator \gsrc, in the sense that they will produce similar outputs given the same latent code $z$.

Crucially, we only regularize the base subspace \wbase, since the subspaces \wi should be allowed to change to learn the new behaviors.
To this end, we project the latent codes to the base subspace \wbase, before calculating the regularization terms.  Our overall regularization objective is thus:
\begin{equation}
\label{eq:regularize_loss}
\begin{split}
    \mathcal{L}_{\text{reg}} = \mathbb{E}_{z \sim p_{\text{src}} (z)} \; \big[ \lambda_{\text{src}}\mathcal{L}_{\text{src}}(G(\prj_{\mathcal{Z}_{\text{base}}}(z))) + \\ \mathcal{L}_{\text{recon}}(G(\prj_{\mathcal{Z}_{\text{base}}}(z))) \big],
\end{split}
\end{equation}
where $\lambda_{\text{src}} = 1$ balances the two terms and $p_{src}(z)$ is the latent distribution over $\mathcal{Z}$ used to train \gsrc.
Our final, regularized domain expansion objective is, therefore:
\begin{equation}
    \mathcal{L}_{\text{full}} = \mathcal{L}_\text{expand} + \mathcal{L}_{\text{reg}}.
\end{equation}

\section{Experiments}
\label{sec:experiments}

We evaluate our method and analyze its key characteristics. \cref{subsec:setting} first details the experimental setting, focusing on StyleGAN2~\cite{karras2020analyzing}. We start by analyzing the knowledge encoded along repurposed directions and compare it to domain adaptation methods (\cref{subsec:effectiveness}). We then delve deeper and evaluate the effects (\cref{subsec:effect_multi}) and opportunities (\cref{subsec:compositionaltiy}) presented by expanding a generator to multiple domains simultaneously.
Next, we demonstrate that the quality of the source domain is maintained in the base subspace (\cref{subsec:no_harm}).
Finally, we demonstrate that our method generalizes to other generators, namely Diffusion Autoencoder~\cite{preechakul2022diffusion}.

Further experiments, results, and details are provided in the supplementary.

\subsection{Experimental Setting}
\label{subsec:setting}

We adopt StyleGAN2 \cite{karras2020analyzing} as the main generator architecture, for its disentangled latent space and because it has been the dominant test bed for generative domain adaptation methods in recent years \cite{gal2021stylegan,ojha2021few,nitzan2022mystyle,bau2020rewriting,zhu2021mind,wang2022rewriting}. 

\myparagraph{Latent space and subspaces.}
Several latent spaces have been considered in the context of StyleGAN. We use the intermediate latent space $\mathcal{W}$ in all our experiments but note it as $\mathcal{Z}$ for consistency.
We use SeFA~\cite{shen2020closedform} for the orthogonal decomposition of $\mathcal{Z}$. 
As SeFA performs SVD, there is a native indication to how dormant is a given latent direction -- the corresponding singular value. As singular values are commonly sorted in decreasing orders, the last basis vectors are most dormant. When expanding with $N$ new domains, unless specified otherwise, we repurpose the last $N$ basis vectors. 
We use $s=20$ in all experiments. 
These decisions are evaluated in greater depth in~\cref{subsec:effect_of_s,subsec:effect_of_dim}.

\myparagraph{Adaptation methods.}

We demonstrate our expansion method with two domain adaptation tasks - StyleGAN-NADA \cite{gal2021stylegan} and MyStyle \cite{nitzan2022mystyle}.
These two tasks were chosen as they differ significantly in key aspects -- source of supervision, sampling distribution and loss.

\texttt{StyleGAN-NADA} is a zero-shot, text-guided, domain adaptation method. It takes as input a pair of text prompts, $t_{\text{source}}$ and $t_{\text{target}}$, describing the desired transformation $\textit{\text{source}} \rightarrow \textit{\text{target}}$ to be applied on the domain of the pretrained generator, \dsrc.
The loss function $\mathcal{L}$ is given by
\begin{equation}
\begin{gathered}
\Delta T = E_{T}\left(t_{\text{target}}\right) - E_{T}\left(t_{\text{source}}\right), \\ 
\Delta I = E_{I}\left(G\left(z\right)\right) - E_{I}\left(\gsrc\left(z\right)\right), \\ 
\mathcal{L} = 1 - \frac{\Delta I \cdot \Delta T}{\left|\Delta I\right|\left|\Delta T\right|}~,
\end{gathered}
\end{equation}\label{eq:directional_loss}
where $E_{I}$ and $E_{T}$ are CLIP's \cite{radford2021learning} image and text encoders respectively.

\texttt{MyStyle} is a few-shot, image-supervised, domain adaptation method. As input, it takes a set of images $\{x_m\}_{m=1}^M$ of an individual ($M \sim 100$), and adapts \gsrc to form a personalized prior for that individual.
The generator is trained to better reconstruct $x_m$ from their original latent space inversions $z_m \in \mathcal{Z}$ \cite{tov2021designing}. Formally, the loss function is given by
\begin{equation}
\begin{split}
    \mathcal{L} = \sum_{m=1}^M \; [ \mathcal{L}_{\text{lpips}}(G(z_m), x_m) +
    \norm{G(z_m) - x_m}_2],
\end{split}    
\end{equation}
where $\mathcal{L}_{\text{lpips}}$ is again the LPIPS loss \cite{zhang2018unreasonable}.

\myparagraph{Datasets and models.}
We demonstrate our method on four datasets -- FFHQ \cite{karras2019style}, AFHQ Dog \cite{choi2020stargan}, LSUN Church \cite{yu2015lsun} and SD-Elephant \cite{mokady2022selfdistilled}. The FFHQ model is expanded with 105 new domains, 100 introduced with the expanded variant of StyleGAN-NADA and 5 from the expanded variant of MyStyle. The AFHQ Dog, LSUN Chruch and SD-Elephant are expanded with 50, 20, and 20 new domains correspondingly, all introduced from the expanded variant of StyleGAN-NADA.

\subsection{Evaluating Domains Individually} 
\label{subsec:effectiveness}
\begin{figure}
    \centering
    \includegraphics[width=\linewidth]{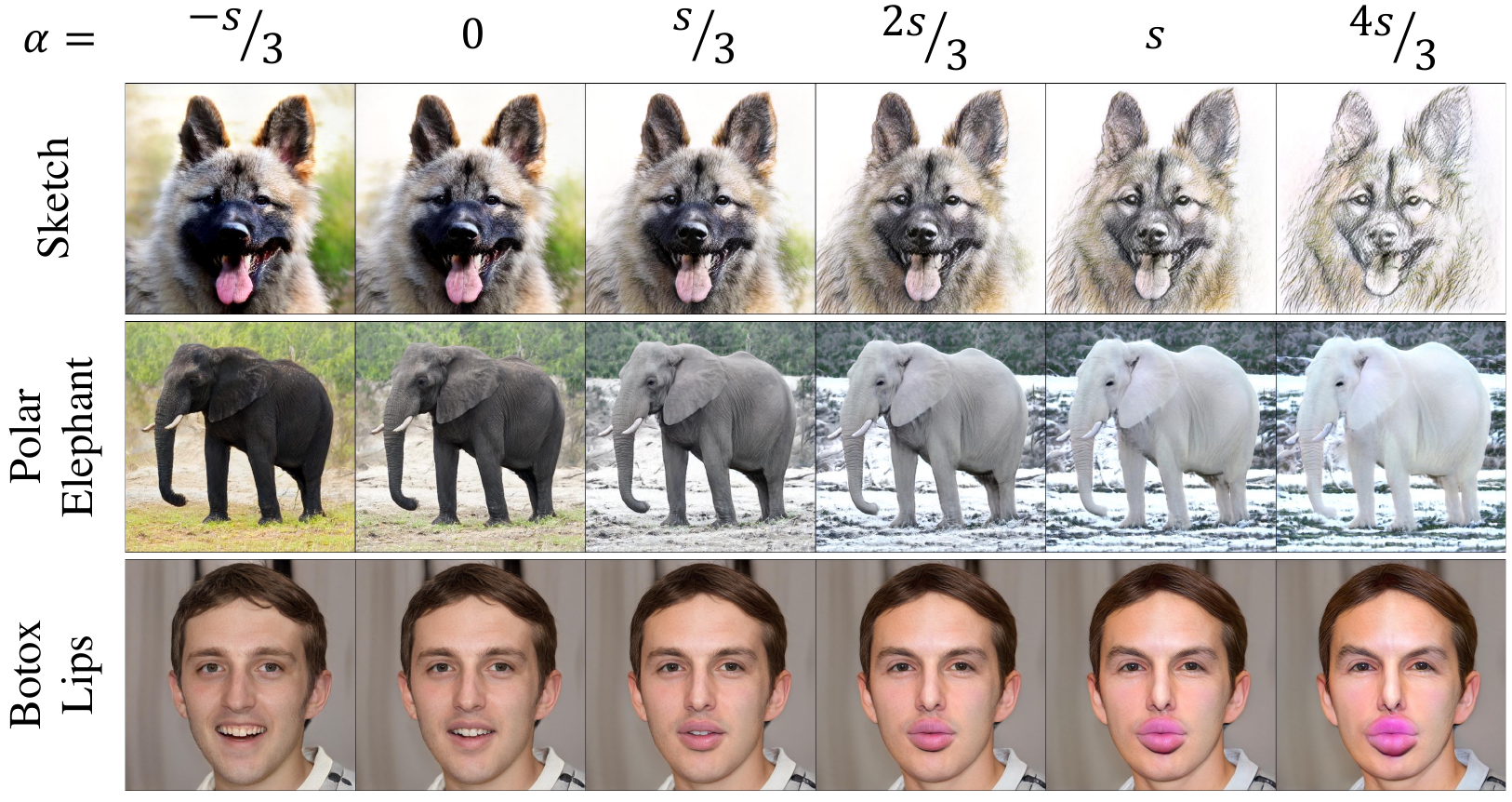}
    \caption{Continuous traversal along repurposed directions.
    As can be seen, the traversal between the base subspace ($\alpha = 0$) and repurposed subspace ($\alpha=s$) portrays a smooth transition between the source and newly introduced domains.
    Advantageously, the semantic meaning of the repurposed direction is preserved in the extrapolation, representing the opposite relationship between the domains $(\alpha < 0)$ or exaggerations of it $(\alpha > s)$.}
    \label{fig:effectiveness_continuous}
\end{figure}

\vspace{3mm}
\myparagraph{Traversing a repurposed direction.}
We start by investigating what knowledge, if any, is encoded along the repurposed latent directions. To this end, starting from a random latent code $z \in \mathcal{Z}_\text{base}$, we individually traverse different repurposed directions, $v_i$, and inspect the generated images $\gt(z + \alpha v_i)$. Sample results from our dogs, elephants, and faces models are displayed in \cref{fig:effectiveness_continuous}.
We find that each individual repurposed direction now successfully encodes the desired factor of variation, in a global and continuous way.

Our training paradigm is inherently discrete -- encouraging the source behavior on the base subspace ($\alpha=0$) and the newly introduced effect on the repurposed space ($\alpha=s$).
Therefore, obtaining a smooth effect might seem surprising at first glance. Nevertheless, this phenomenon can be clearly traced to the well-established observation that generators are smooth with respect to their latent space \cite{karras2019style}. 

\begin{figure}
    \centering
    \includegraphics[width=\linewidth]{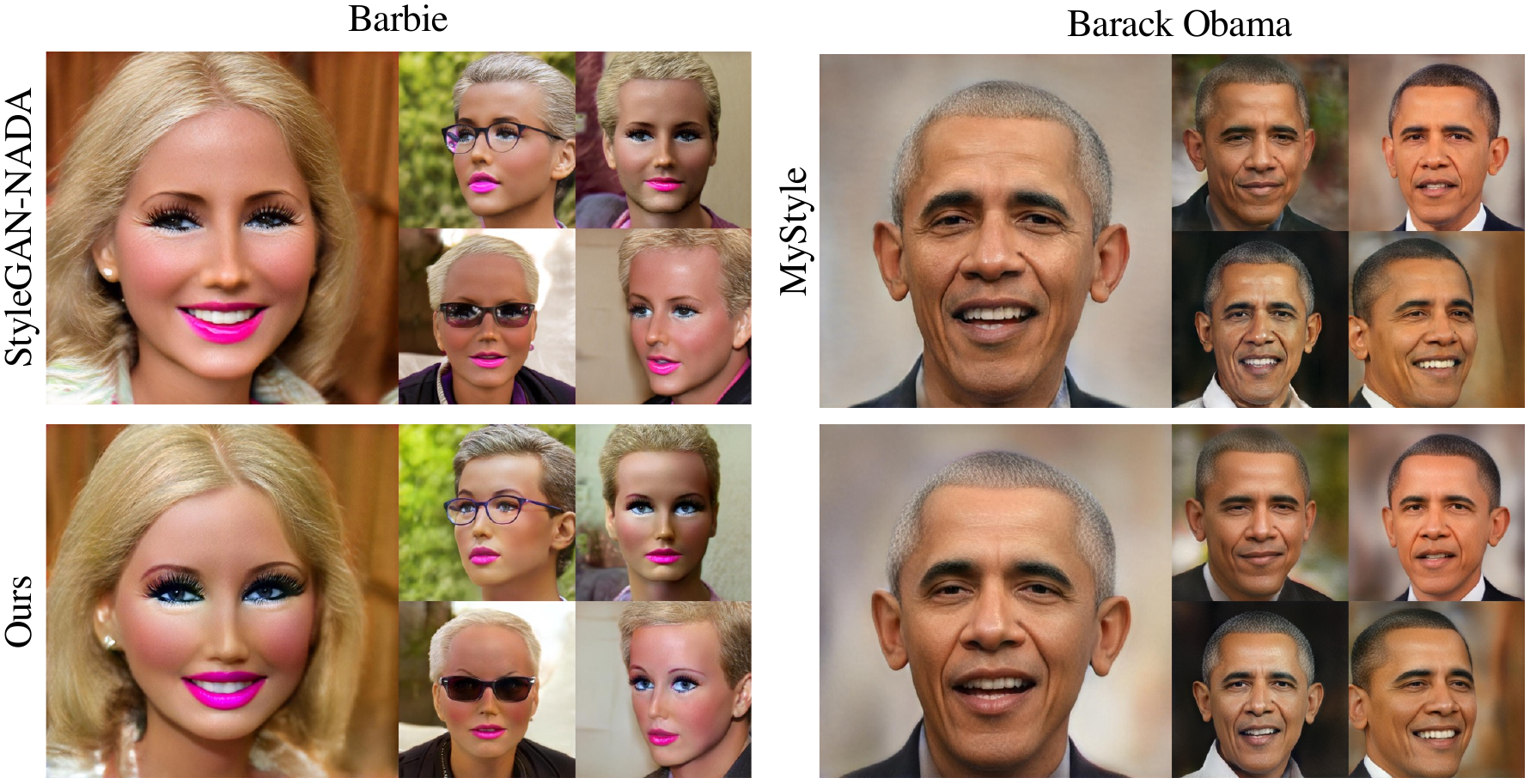}
    \caption{A random set of images generated by our generator from repurposed subspaces (bottom) and by corresponding domain adaptation methods (top). The images are similar and differences are subtle.}
    \vspace{-5mm}
    \label{fig:effectiveness_compare}
\end{figure}

\begin{table}
    \centering
    \setlength{\tabcolsep}{1.5pt}
    \begin{tabular}{lccc}
         \toprule
         Method & User \% $(\uparrow)$ & ID $(\uparrow)$ & Diversity\small{$\times 10$} $(\uparrow)$ \\
         \midrule
         StyleGAN-NADA & 41.2\% & - & 2.42 $\pm$ 0.13  \\
         Ours w/ NADA & \textbf{58.8}\% & - & 2.42 $\pm$ 0.13 \\
         \midrule
         MyStyle & - & \textbf{0.80 $\pm$ 0.06} & 3.08 $\pm$ 0.15 \\
         Ours w/ MyStyle & - & 0.76 $\pm$ 0.05 & \textbf{3.14 $\pm$ 0.14} \\
         \bottomrule
    \end{tabular}
    \caption{Quantitative comparison of images generated from our repurposed subspaces to those generated by corresponding domain adaptation methods - StyleGAN-NADA \cite{gal2021stylegan} and MyStyle \cite{nitzan2022mystyle}. We follow each adaptation method's quantitative evaluation protocol.}
    \label{tab:effective_quant}
\end{table}

\myparagraph{Behavior on the repurposed subspace.}
We have transformed adaptation tasks into expansion tasks by limiting the training effect to the repurposed subspaces only. But, for latents in repurposed subspaces ($\alpha = s$), the domain adaptation could be considered to have been applied as-is. 

We next directly compare the images generated by our generator from the repurposed subspace to the corresponding images generated by the domain-adapted generator. 
We inherit and repeat the quantitative evaluation protocols performed by each of the adaption tasks.
To compare quality with StyleGAN-NADA \cite{gal2021stylegan} we perform a two-alternative forced choice user study. Users were asked to pick the image that has higher-quality and better aligns with the target text used for training. We gathered $1440$ responses from $32$ unique users. 
To compare quality with MyStyle~\cite{nitzan2022mystyle}, we evaluate preservation of identity in generated images, as observed by a face recognition network~\cite{huang2020curricularface}.
For both methods, the diversity is compared based on intra-cluster LPIPS \cite{zhang2018unreasonable} distance, first suggested by Ojha \etal \cite{ojha2021few}.
We use $10$ domains for comparison with StyleGAN-NADA and $5$ for comparison with MyStyle. Note that we use a single generator \gt, expanded with $105$ domains, while competing methods use a dedicated model per domain, $15$, overall.
We report the results in \cref{tab:effective_quant,fig:effectiveness_compare}.

As can be seen, on the repurposed subspaces, our method produces comparable images to that generated by the dedicated, domain-adapted generator. Perhaps surprisingly, users somewhat prefer our results over StyleGAN-NADA's. We speculate this is due to the significantly greater difficulty of choosing hyperparameters for their training. 

\begin{figure}
    \centering
    \begin{subfigure}{\linewidth}
         \centering
         \includegraphics[width=\linewidth]{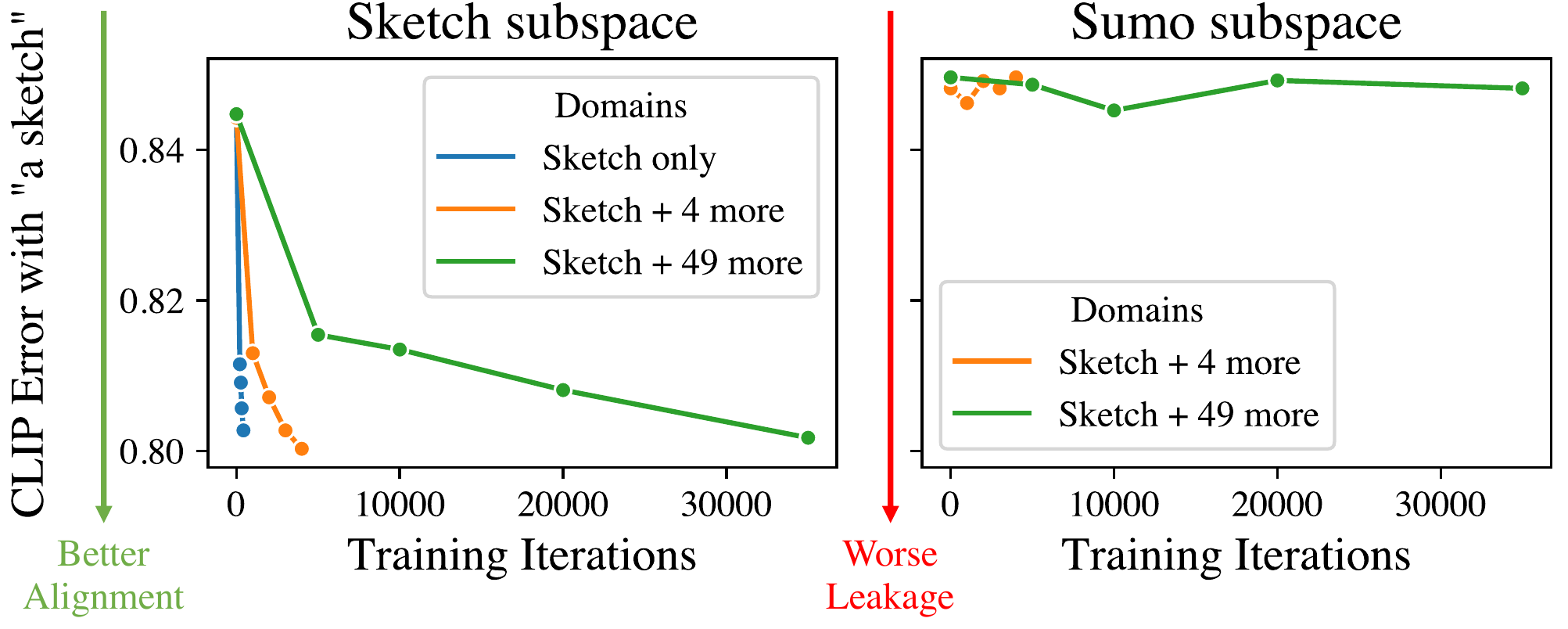}
         \caption{}
         \label{fig:multi_quant}
    \end{subfigure}
    \hfill
    \vspace{1mm}
    \begin{subfigure}{\linewidth}
         \centering
         \includegraphics[width=\linewidth]{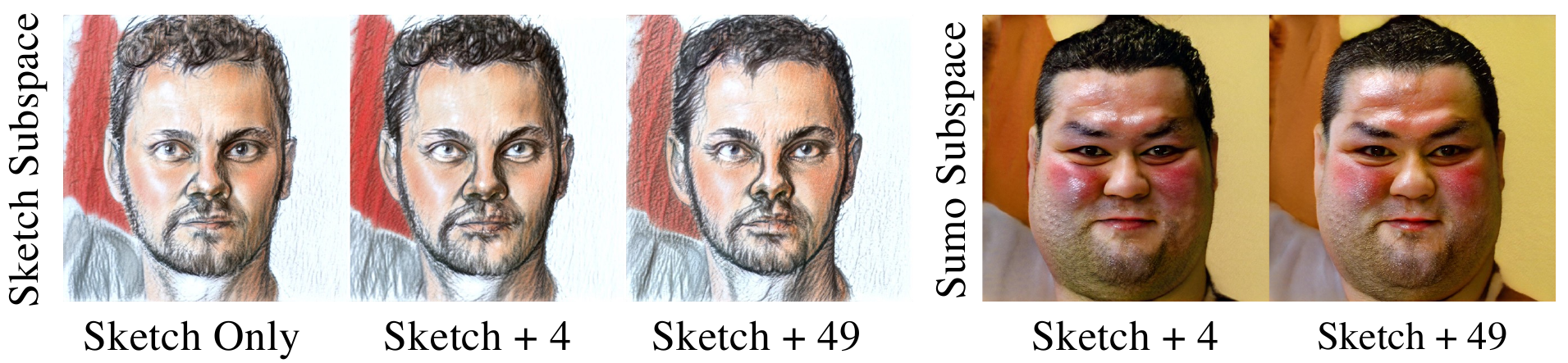}
         \caption{}
         \label{fig:multi_qual}
    \end{subfigure}
    \caption{Investigating the effect of introducing multiple domains simultaneously. 
    (a) Reports the CLIP error of generated images with the text ``a sketch'', as a function of training iterations. Images are generated from the ``Sketch'' and ``Sumo'' subspaces of models trained with a different number of domains.
    (b) Depicts generated images from models that have similar CLIP errors.
    As can be seen, the sketch domain does not ``leak'' into the sumo subspace. Additionally, introducing additional domains delays, but does not prevent, the introduction of sketch. 
    }
    \label{fig:multi}
\end{figure}
\begin{figure*}
    \centering
    \includegraphics[width=0.97\linewidth]{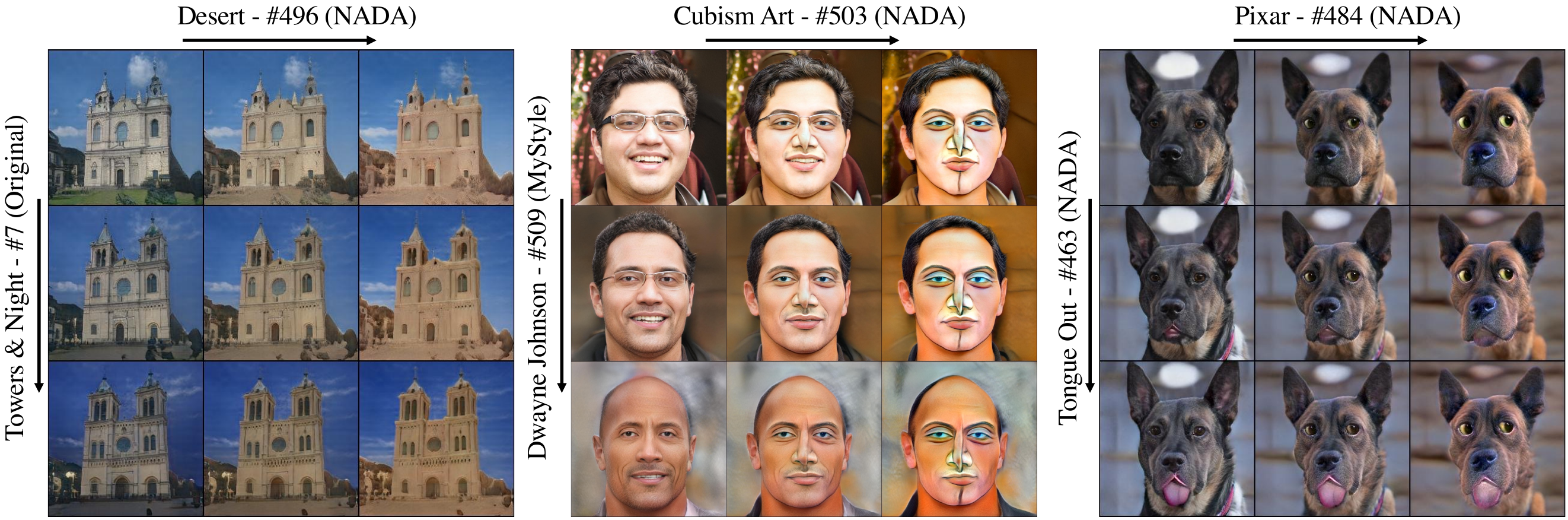}
    \caption{Composing multiple effects by simple latent traversal. In each grid, we start from the latent code that generates the top-left image and traverse along two latent directions, represented by advancement in rows and columns. For each direction, we note the associated domain, its ordinal number in the latent space's basis, and the training method used (``NADA'' or ``MyStyle'') to learn the domain. As can be seen, \gt has learned a disentangled representation, allowing meaningful composition of concepts. Specifically, note the disentanglement between directions, as traversing left-right does not affect the magnitude of the effect corresponding to up-down traversal, and vice versa.}
    \vspace{-2mm}
    \label{fig:composition_showcase}
\end{figure*}

\subsection{Effect of Domains on Each Other}
\label{subsec:effect_multi}

Previous evaluation of individual repurposed directions already indicates disentanglement between different factors of variation. For example, ``Barbie'' images in \cref{fig:effectiveness_compare} show no sign of being caricature, Barack Obama, or any of the other hundred factors of variation introduced to that generator.
In this section, we delve deeper into evaluating the effects of expanding with multiple factors of variation.

To this end, we train three models to expand the FFHQ parent model with either 1, 5 or 50 new domains, all induced by StyleGAN-NADA \cite{gal2021stylegan}. All models are expanded with ``Sketch'', and the latter two with ``Sumo'' as well as other factors of variation.
We quantify the strength of introduced factor of variation using \textit{CLIP error}, the 1-complement of the score produced by CLIP \cite{radford2021learning}. We use the top-performing version of the CLIP encoder available, ViT-L/14, which is not used during training. We note that simply minimizing CLIP error is not the objective, as it might lead to favoring mode-collapsed and adversarial examples~\cite{gal2021stylegan}. Nevertheless, together with qualitative inspection, it is useful for comparing different models.

In \cref{fig:multi_quant}, we report the CLIP error of generated images with the text ``a sketch'', as a function of the training iterations. Images are generated from the ``Sketch'', and if exists, ``Sumo'' subspace.
First, we observe that CLIP error is decreasing for ``Sketch'' subspaces in all models, as expected.
Conversely, CLIP error in the ``Sumo'' subspace does not significantly change, indicating it is not becoming any more or less of a sketch. This result quantitatively supports our previous finding, that factors of variations do not interfere with each other, and demonstrates it is true regardless of the number of other factors of variations learned simultaneously. 
Additionally, we observe that expanding with additional factors of variation delays, but does not prevent, \gt the introduction of ``Sketch'' effectively. The observed delay is expected, as expanding with more variations corresponds to \gt optimizing and balancing additional loss functions.
Generated samples from the sketch and sumo subspaces are provided in \cref{fig:multi_qual}.

\subsection{Compositionality}
\label{subsec:compositionaltiy}
While accidental ``leakage'' between latent directions during training is undesired, intentionally composing variations at test time is useful.
For generative models with a disentangled latent space, summing together latent directions aggregates their semantic meaning, and should not affect the magnitude of their effects if applied separately. For example, if direction $v_1$ controls head pose and direction $v_2$ controls an unrelated variation, images $G(z + v_1 + v_2)$ and $G(z + v_1)$ should depict the same head pose.

We find that the latent space of the expanded generator \gt is disentangled, and variations can indeed be composed effectively. Crucially, variations can be composed with each other regardless of their originating training task, including those on the base subspace, learned from the source domain. \cref{fig:composition_showcase} shows a sample of gradual composition results across models and training tasks.

\myparagraph{Comparison to existing techniques.}
Several domain adaptation methods have proposed techniques to combine multiple variations. These methods still train a {\em separate} generative model per variation, but combine their effects in test-time.
Specifically, in the realm of CLIP-supervised training, StyleGAN-NADA \cite{gal2021stylegan} interpolates the generators' weights, while DiffusionCLIP \cite{kim2022diffusionclip} interpolates intermediate activations of the generators.
Next, we compare the disentanglement of composition in our generator to that made possible using these techniques.

\begin{figure}
    \centering
    \vspace{-5mm}
    \begin{subfigure}{\linewidth}
         \centering
         \includegraphics[width=\linewidth]{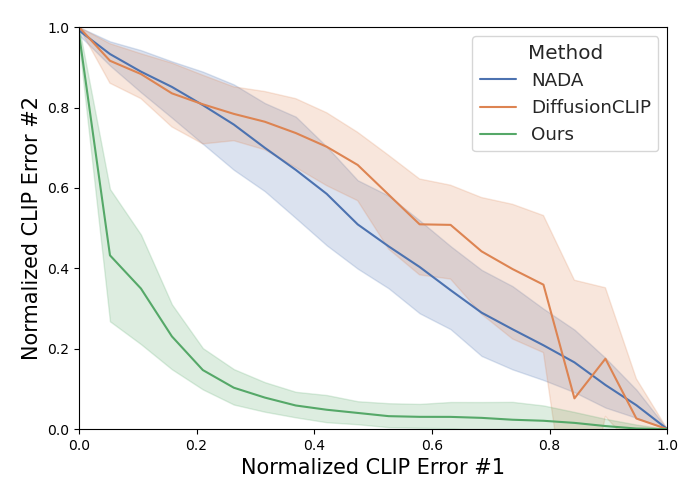}
         \vspace{-6mm}
         \caption{}
         \label{fig:composition_quant}
    \end{subfigure}
    \hfill
    \vspace{1mm}
    \begin{subfigure}{\linewidth}
         \centering
         \includegraphics[width=\linewidth]{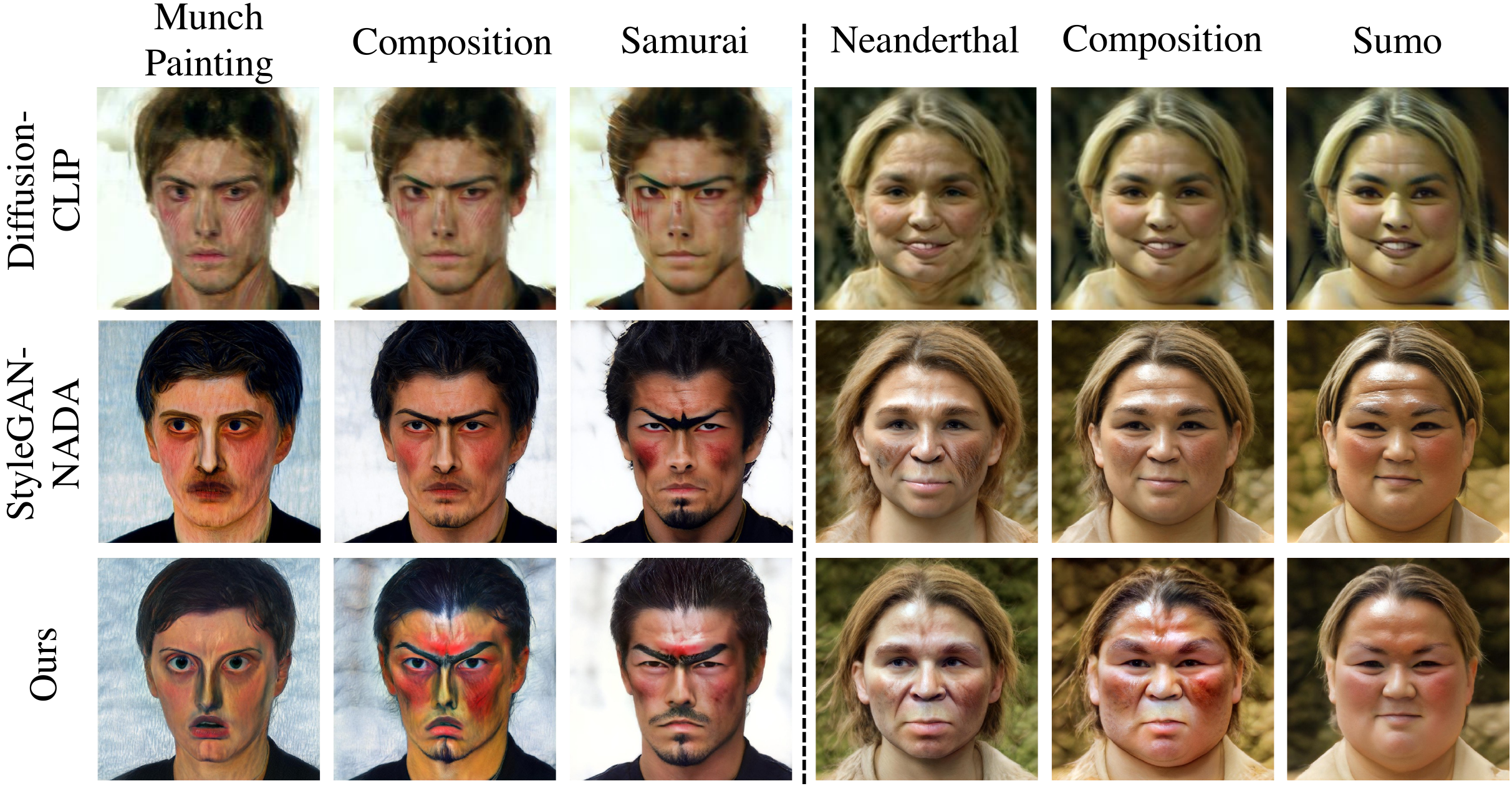}
         \caption{}
         \label{fig:composition_qual}
    \end{subfigure}
    \vspace{-2mm}
    \caption{Comparing compositionality in our generator to methods of combining multiple domains proposed by StyleGAN-NADA \cite{gal2021stylegan} and DiffusionCLIP \cite{kim2022diffusionclip}. Starting from a setting optimized for either text prompt \#1 or \#2, we gradually introduce the variation described by the other text. (a) Reports the CLIP error to both prompts along the gradual introduction, normalized to the error obtained for each text prompt in isolation. (b) Portrays a sample of qualitative results, where the composition is such that assigns equal strengths to both effects. As can be seen in, both quantitatively and qualitatively, NADA and DiffusionCLIP directly tradeoff one effect for the other -- strengthening the effect of one prompt directly lessens that of the other. In contrast, our generator allows true composition of modalities.
    }
    \label{fig:composition_comparison}
\end{figure}

For each method, we start with a setting that was optimized to generate images that align with one of two text prompts. In our case, this setting is \gt with latent codes in certain subspaces. For the baselines, these settings are dedicated generators with any latent code. Then, for each method, we gradually introduce the variation described by the other text prompt and generate the corresponding images. Finally, we measure normalized CLIP error between generated images and the two prompts. We normalize all errors by the error of the initial setting, to make the metric comparable across methods and text prompts. \cref{fig:composition_comparison} reports the mean and standard deviation of the CLIP error, on $10$ pairs of prompts, and provides a sample of qualitative results.
As can be seen, both baseline methods directly tradeoff one domain for the other, expressed by a linear-looking trend. Conversely, our method obtains significantly lower errors and allows for a true composition of concepts. 

\subsection{Preservation of the Source Domain}
\label{subsec:no_harm}

\begin{table}
    \centering
    \setlength{\tabcolsep}{5pt}
    \begin{tabular}{lcccc}
         \toprule
         Model & FFHQ & AFHQ & \shortstack{LSUN \\ Church} & \shortstack{SD\\Elephant} \\
         \midrule
         Parent & 2.77 & 7.43 & 3.92 & 2.30 \\
         Ours  & 2.80 & 7.51 & 3.76 & 2.70 \\ 
         \midrule
         \small{Only $\mathcal{L}_\text{src}$} & \small{2.75$\pm$0.08} & \small{7.38$\pm$0.09} & \small{3.31$\pm$0.22} & \small{3.91$\pm$0.67} \\
         \bottomrule
    \end{tabular}
    \caption{We generate images from the base subspace and report FID \cite{heusel2017gans} $(\downarrow)$ with respect to source domain dataset. We compare our FID to that of the source generator $G_\text{src}$. For reference, we also continue training the source generator for the same number of iterations with its original loss - $\mathcal{L}_\text{src}$, and report the mean and standard deviations of FID along the training. As can be seen, on the base subspace, our models have comparable FID scores to their parents. Furthermore, similar magnitude of change in FID are observed by simply continuing training, indicating that the change in FID might be, at least in part, due to ``random'' fluctuations.}
    \label{tab:no_harm_base_hyperplane}
\end{table}

We next evaluate the preservation of the source domain in \gt.
To this end, using FID~\cite{heusel2017gans}, we compare the quality of images generated from the base subspace \wbase of \gt to those generated by the source generator $G_{src}$.
Since the generator is being trained, some change in FID is expected.
Therefore, we also report the average and standard deviation over FID scores for a generator that simply continues training, \ie, using only the original loss \lsrc.
Results are reported in \cref{tab:no_harm_base_hyperplane} and vary between datasets.
\yn{
Across all datasets, we observe that the FID from our base subspace is within $1\sigma$ of the that obtained from either the parent or a generator that continues training.
For reference, generator's adapted with StyleGAN-NADA and MyStyle have an average FID of $125$ and $183$, respectively.
}
We conclude that the expansion method might have a slight impact on FID, but it is negligible. 

\subsection{Generalization Beyond StyleGAN}
\label{subsec:diffae}
We finally demonstrate that our expansion method generalizes to additional models, by experimenting with Diffusion Autoencoder (DiffAE) ~\cite{preechakul2022diffusion}.

\begin{figure}[h]
    \centering
    \includegraphics[width=0.95\linewidth]{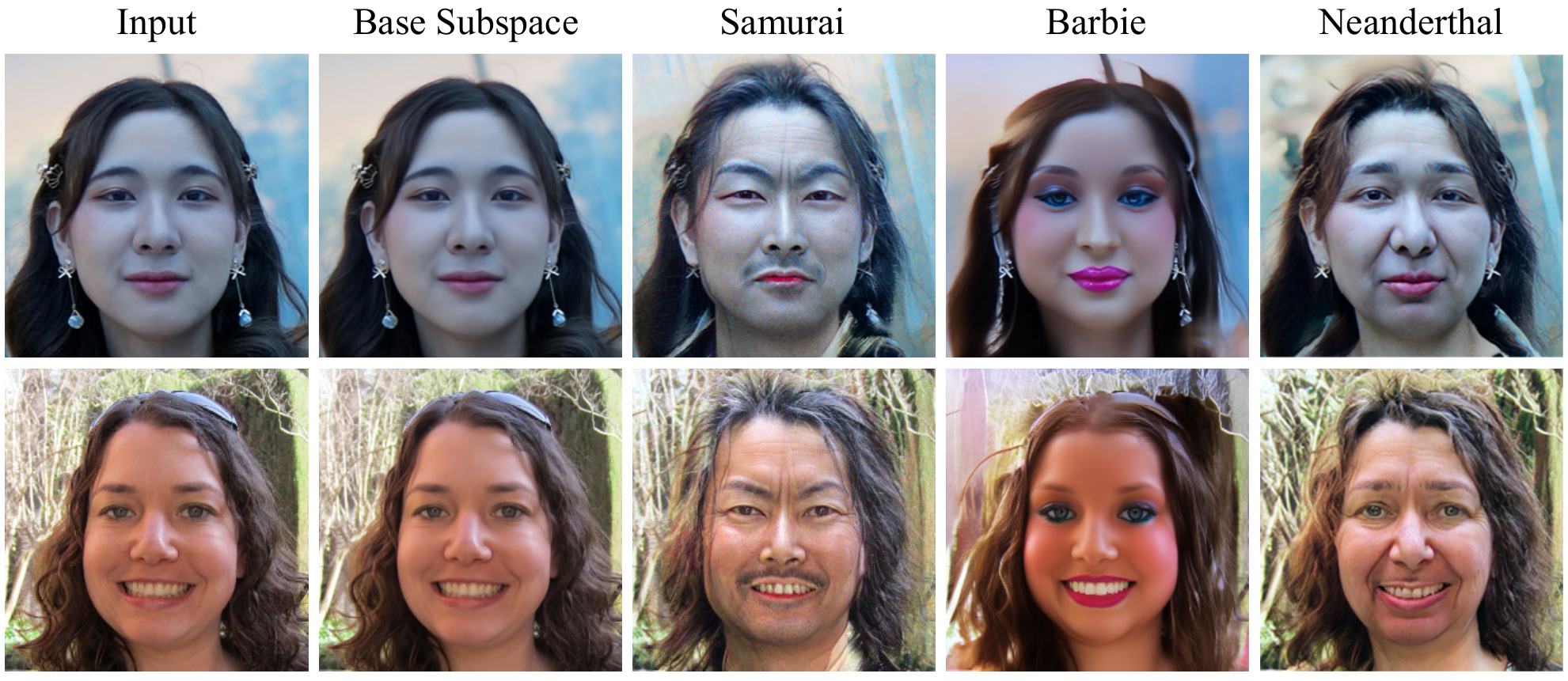}
    \caption{
    Applying our domain expansion method with Diffusion Autoencoder~\cite{preechakul2022diffusion}.
    } 
    \label{fig:diffae}
    \vspace{-3mm}
\end{figure}

DiffAE differs from StyleGAN significantly in both architecture and training method.
Nevertheless, it similarly possesses a semantic latent space (dubbed $z_{\text{sem}}$), which is the only prerequisite of our expansion method.
We identically apply our expansion method to DiffAE, making no modifications.
Specifically, we decompose the latent space $z_{\text{sem}}$ using SeFA~\cite{shen2020closedform}, and repurpose the dormant directions with StyleGAN-NADA~\cite{gal2021stylegan} induced domains.
We find that our expansion method works just as well with DiffAE, and provide a sample of qualitative results in~\cref{fig:diffae}.

\section{Conclusions}

We present a new problem -- \emph{domain expansion} -- and propose an approach to solve it.
The core of our method is to carefully structure the latent space, such that it is amenable to learning additional knowledge, while keeping the existing knowledge intact. 
Our method takes advantage of the existence of dormant latent directions, and the task itself implicitly relies on the capacity of the model weights to represent more knowledge. If one of these assumptions does not hold, it might not be possible to apply domain expansion. 
However, the popularity of methods squeezing neural networks, such as Knowledge Distillation~\cite{hinton2015distilling}, and current estimates of the intrinsic dimensionality of image datasets~\cite{pope2021intrinsic}, indicate that these assumptions commonly hold.
In our experiments, we were able to expand to hundreds of directions. A plausible limitation is that the model can be expanded to a certain point but ultimately limited by factors such as the latent space or network capacity. Overcoming this limitation, perhaps by considering more complex latent space structures, is an avenue for future work.

\myparagraph{Acknowledgment.}
We are grateful to Rinon Gal, Yossi Gandelsman and Sheng-Yu Wang for their suggestions in the early stages of this research. We also thank Rinon Gal, Yossi Gandelsman, Kfir Aberman, Alon Nitzan, Omer Bar-Tal, Nupur Kumari and Gaurav Parmar for proofreading the draft. We also thank Nupur Kumari for a technical advice and to Asaf Weinberg and Daniel Garibi for help running the Diffusion Autoencoder experiment. We finally thank Yogev Nitzan for his help with the user study and for coming up with hundreds of textual prompts.
This work was done while Yotam Nitzan was an intern at Adobe.
This research was supported in part by the Israel Science Foundation (grants no. 2492/20 and 3441/21), Len Blavatnik and the Blavatnik family foundation, and The Tel Aviv University Innovation Laboratories (TILabs).

{\small
\bibliographystyle{ieee_fullname}
\bibliography{main}
}

\newpage

\section*{Appendices}

\appendix
\section{Appendices Overview}

In \cref{sec:cgan_baseline}, we consider a baseline for domain expansion and demonstrate it is inferior to our proposed method. 
Next follows the main part of the supplementary, \cref{sec:supp_exp}, in which we perform additional analysis and experimentation of our method.
Finally, in \cref{sec:supp_details}, we provide additional details completing the paper.

\section{Domain Expansion Baseline Using Class-Conditioning}
\label{sec:cgan_baseline}
\begin{figure*}
    \centering
    \begin{subfigure}[t]{0.5\linewidth}
        \centering
        \includegraphics[width=0.97\linewidth]{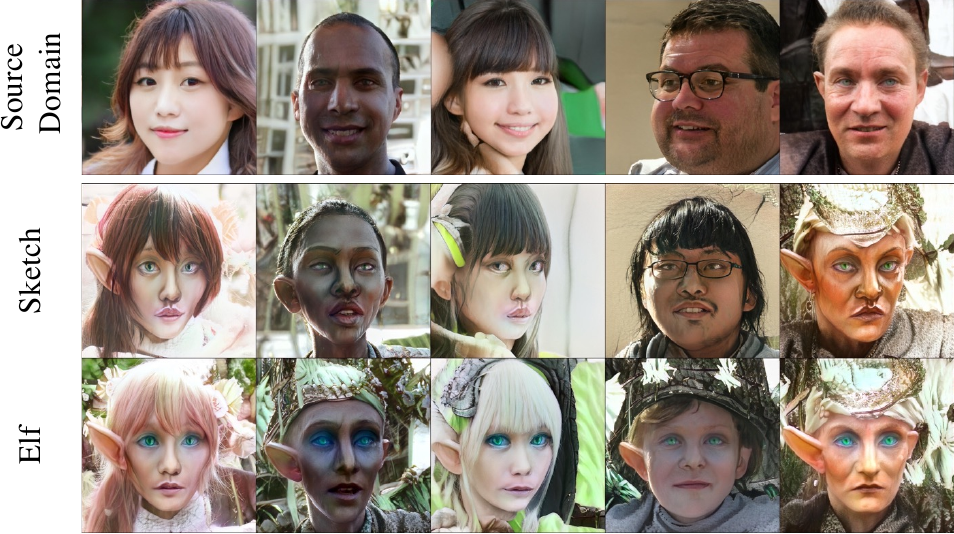}
        \caption{Class-conditioned baseline}
        \label{fig:cgan_cgan}
    \end{subfigure}%
    ~
    \begin{subfigure}[t]{0.5\linewidth}
        \centering
        \includegraphics[width=0.97\linewidth]{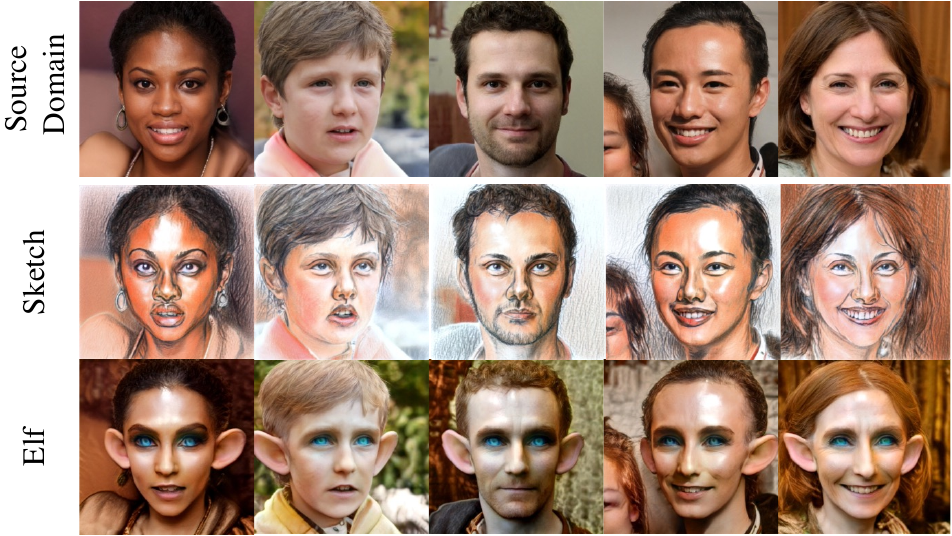}
        \caption{Our domain expansion method}
        \label{fig:cgan_ours}
    \end{subfigure}
    
    \caption{Experimenting with a class-conditioned baseline for domain expansion. (a) Images generated from a class-conditioned expanded model from the same $z$ latent codes for the source, sketch, and elf domains. The source domain is preserved well in its dedicated class. However, the newly introduced domains ``leak'' information, expressed in long, elf-like, ears in the sketch domain. Additionally, the different domains are not well-aligned, as changing the domain also results in unrelated changes to head pose and facial expressions. (b) Comparable results from our domain expansion method, provided for reference. As can be seen, using our method, the domains do not interfere with each other and are well-aligned.}
    
    \label{fig:cgan}
\end{figure*}

In this section, we experiment with an alternative, baseline, method to perform domain expansion.
Generative models capturing multiple domains commonly use a class-conditioning mechanism~\cite{brock2018large}.
Adopting this approach, we attempt to perform domain expansion by modeling domains with classes. We find that this method does not work as well as our proposed method.

\myparagraph{Method.} 
We start with an unconditional pretrained generator, specifically StyleGAN \cite{karras2020analyzing}. We then make the generator condition on a one-hot vector, using the architecture proposed by Karras \etal \cite{Karras2020ada}. This change involves adding a single MLP layer, whose input is the one-hot vector. Its output is concatenated to the random latent code and then fed to the generator.

The class-conditioned generator is trained in a similar protocol to our method.
The source domain uses class $c=0$, which is analogous to the base subspace. 
Whenever the $0^\text{th}$ class is sampled, we apply the original loss \lsrc and the \textit{memory replay} regularization (See \cref{subsec:regularization}). 
Formally, the loss describing this training is
\begin{equation}
\begin{split}
    \mathcal{L}_{\text{reg}} = \mathbb{E}_{z \sim p_{\text{src}} (z)} \; \big[ \lambda_{\text{src}}\mathcal{L}_{\text{src}}(G(z, c=0)) + \\ \mathcal{L}_{\text{recon}}(G(z, c=0)) \big],
\end{split}
\end{equation}
where $\mathcal{L}_{\text{recon}}$ is the memory-replay loss defined in \cref{eq:recon_loss} and $\lambda_{\text{src}}=1$ is a hyperparameter weighting the losses.
Other classes, analogous to repurposed subspaces, are dedicated to the newly introduced domains.
Whenever the $i^\text{th}$ class is sampled ($i > 0$), we apply the loss of the domain adaptation task \li.
Applied over all new domains, the expansion loss is formally given by
\begin{equation}
    \mathcal{L}_{\text{expand}} = \sum_{i=1}^N \; \mathbb{E}_{z \sim p_i (z)} \;
    \mathcal{L}_i(G(z, c=i)).
\end{equation}
The final training objective still reads as $\mathcal{L}_{\text{full}} = \mathcal{L}_{\text{expand}} + \mathcal{L}_{\text{reg}}$.

\myparagraph{Experiments.}
We expand an FFHQ \cite{karras2019style} generator with two new domains, ``Sketch'' and ``Tolkien Elf'', introduced using StyleGAN-NADA \cite{gal2021stylegan}. We display the
generated images using the same $z$ latent codes for the different classes \cref{fig:cgan_cgan}. 

We qualitatively observe that the expanded, class-conditioned generator preserves the source domain well, also expressed by preserving the FID \cite{heusel2017gans} score. 
However, for new domains, we observe degraded performance from two aspects. First, the class-conditioned generator ``leaks'' knowledge between the classes. For example, in \cref{fig:cgan_cgan}, faces generated from the class dedicated to sketches also have long, elf-like, ears.
Second, the domains are not ``aligned''. Despite being generated from the same $z$ latent codes, the images differ beyond the differences between domains. For example, corresponding images from the source domain and elf domain often portray different head poses and facial expression. Therefore, it is not clear how can one obtain the elf ``version'' of a given face image, limiting the applications of such a model.

For reference, we display comparable results from our expansion method in \cref{fig:cgan_ours}. As can be seen, our method does not suffer from these issues.

\section{Additional Experiments}
\label{sec:supp_exp}

\subsection{Latent Directions Analysis}
\label{subsec:analysis}
\begin{figure}
    \centering
    \begin{subfigure}{\linewidth}
        \centering
        \includegraphics[width=\linewidth]{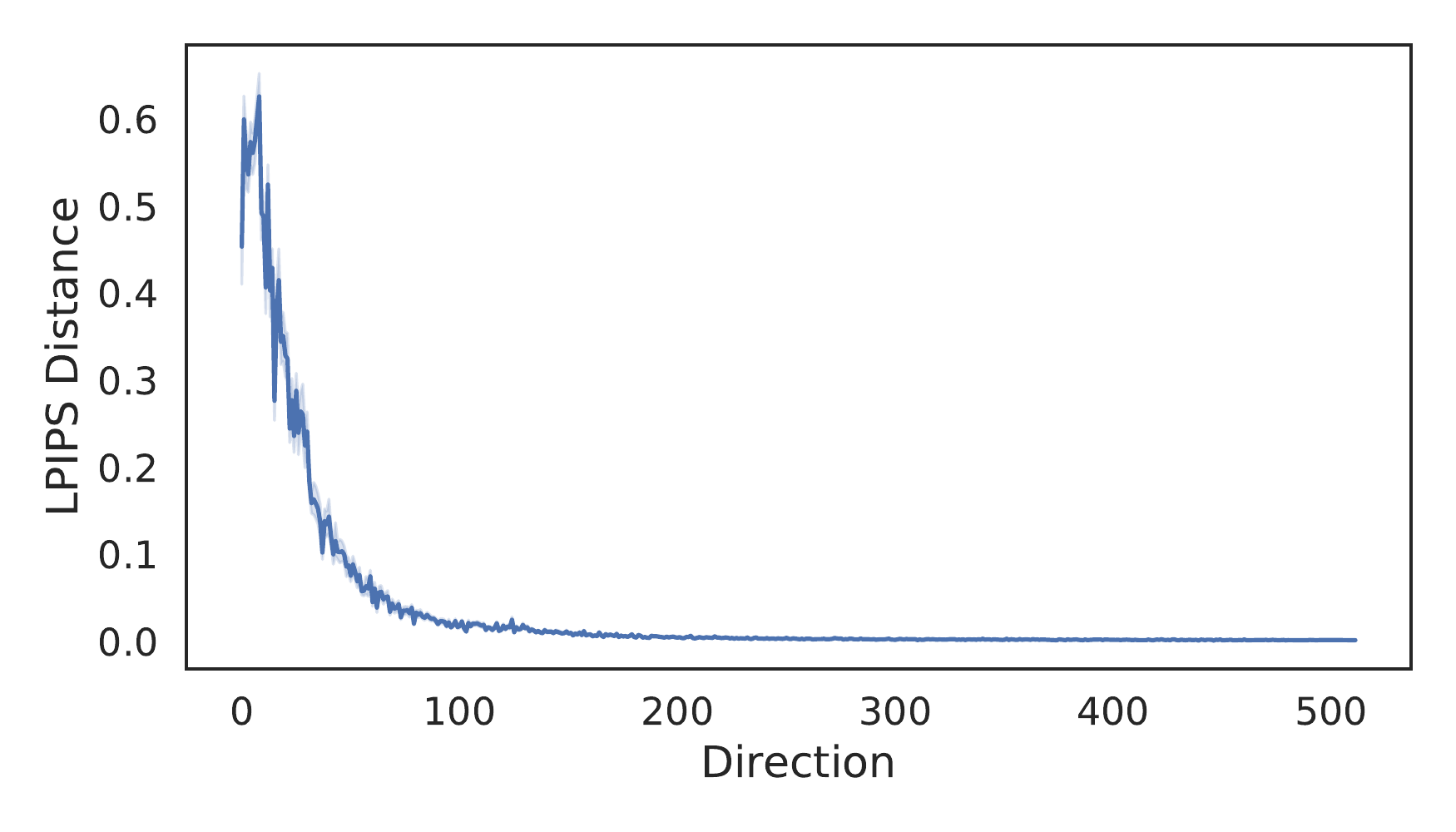}
        \caption{FFHQ}
        \label{fig:ffhq_lpips_per_dim}
    \end{subfigure}
    \vspace{1mm}
    \begin{subfigure}{\linewidth}
        \centering
        \includegraphics[width=\linewidth]{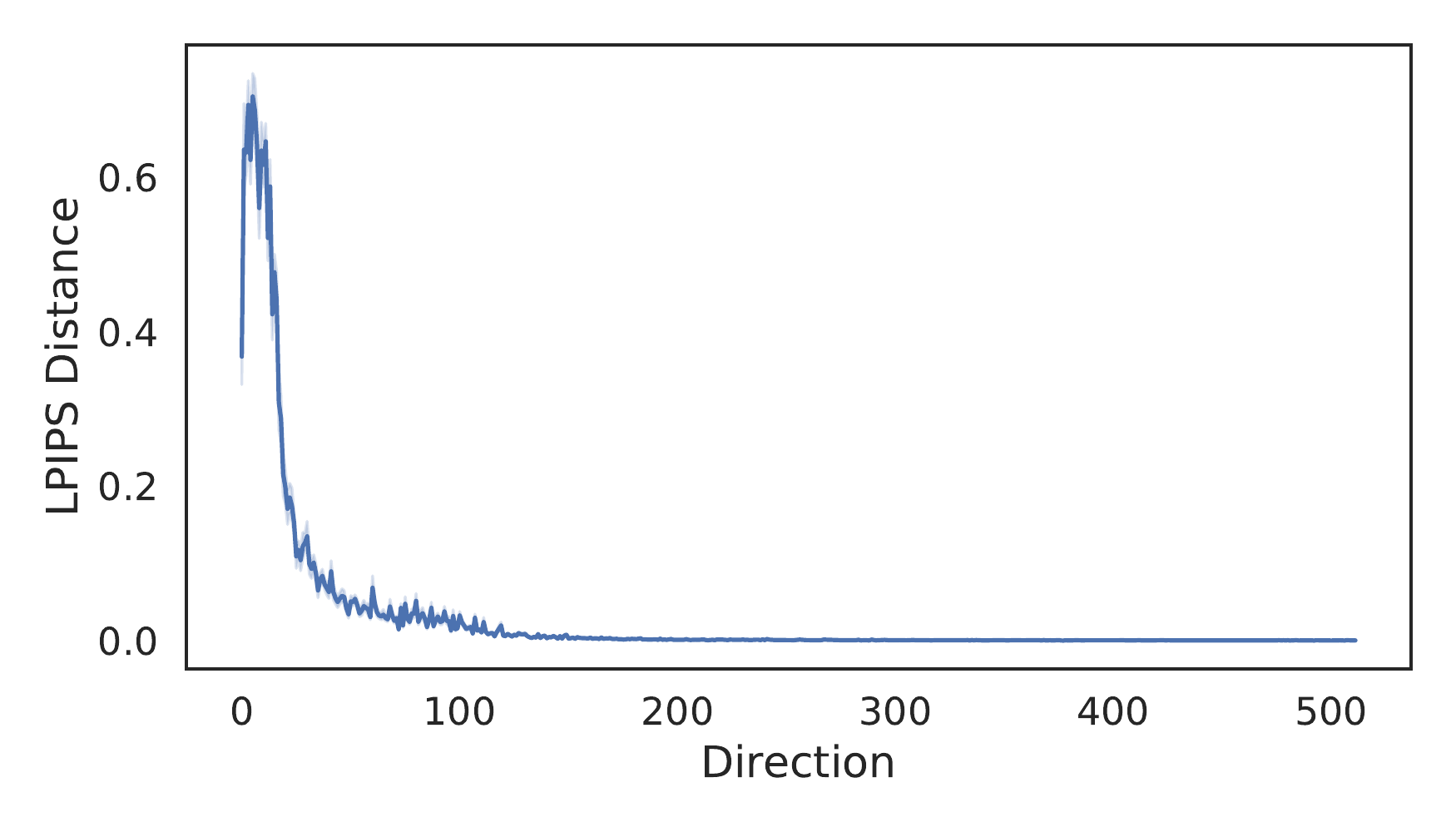}
        \caption{LSUN Church}
        \label{fig:church_lpips_per_dim}
    \end{subfigure}
    \caption{Magnitude of perceptual effect caused by traversing different directions. 
    Directions are sorted in decreasing order according to their corresponding singular values.
    For each direction, we measure the LPIPS distance \cite{zhang2018unreasonable} between images from two latent codes distanced by a $3\sigma$ traversal along the direction. 
    As can be seen, the effect caused by the traversal diminishes quickly and the majority of directions are dormant.
    }
    \label{fig:lpips_per_dim}
\end{figure}

\begin{figure*}
    \centering
    \begin{subfigure}[t]{0.5\linewidth}
        \centering
        \includegraphics[width=0.64\linewidth]{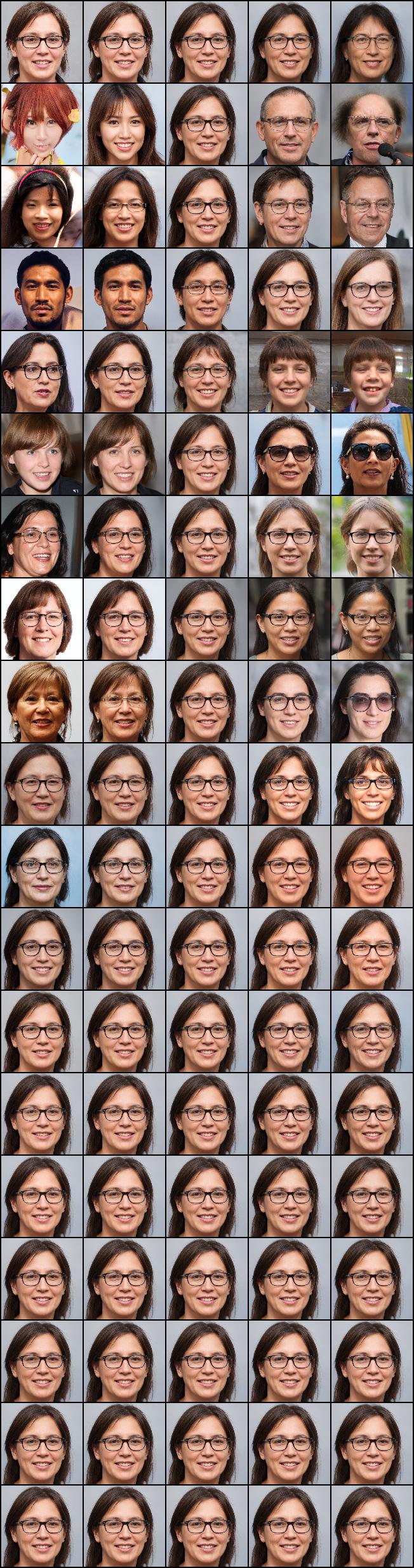}
        \caption{FFHQ}
        \label{fig:ffhq_synth_per_dim}
    \end{subfigure}%
    ~
    \begin{subfigure}[t]{0.5\linewidth}
        \centering
        \includegraphics[width=0.64\linewidth]{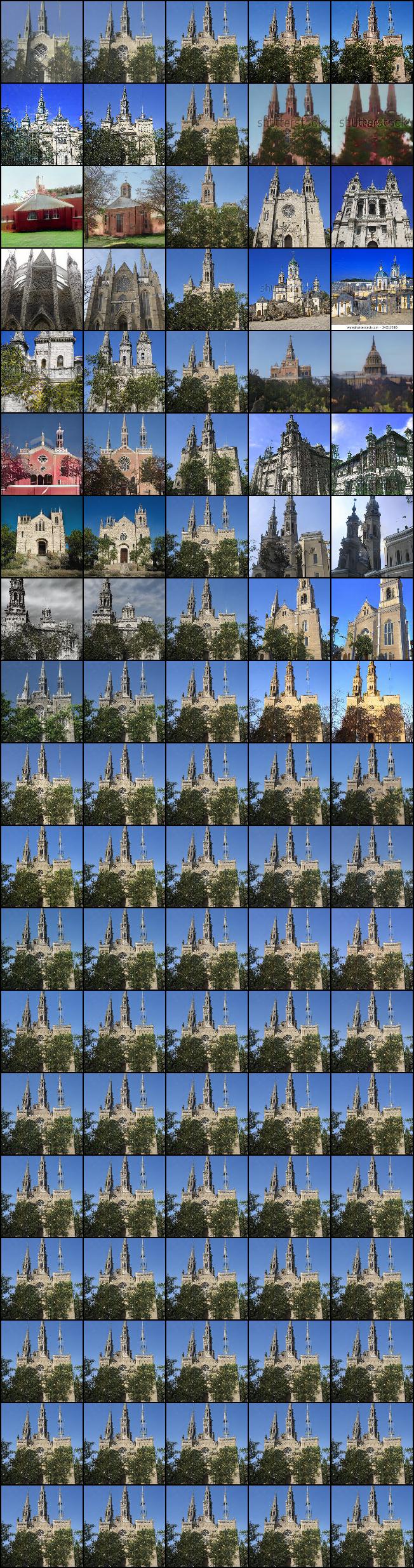}
        \caption{LSUN Church}
        \label{fig:church_synth_per_dim}
    \end{subfigure}
    \caption{Visualization of $\pm 3\sigma$ traversal along latent directions in the FFHQ \cite{karras2019style} and LSUN Church \cite{yu2015lsun} models, obtained using SeFA \cite{shen2020closedform}. Directions shown are sorted from least ($v_0$, top) to most ($v_{511}$, bottom) dormant. As can be seen, later directions are dormant -- not affecting the generated image. We over-sample early directions for clarity. In practice, over 80\% of directions are dormant.}
    \label{fig:synth_per_dim}
\end{figure*}

Our method explicitly relies on the existence of dormant directions and their distinction from non-dormant directions.
We wish to emphasize that the dichotomous distinction between ``dormant'' and ``non-dormant'' is a simplification.
In \cref{fig:lpips_per_dim}, we report the mean LPIPS distance induced to images by a $3\sigma$ traversal along each direction. As can be seen, the distance is never exactly $0$ and there is also no clear discontinuity. 
Nevertheless, it is clear that later directions, usually those beyond $100$, cause significantly smaller perceptual change in the generated image. This behavior can also be qualitatively observed in \cref{fig:synth_per_dim}.

As discussed in \cref{subsec:setting}, this ``almost'' monotonous behavior is expected as our latent directions are right-singular vectors, sorted in decreasing order according to their corresponding singular values \cite{shen2020closedform}.

\subsection{Effect of Choice of Direction for Domain}
\label{subsec:effect_of_dim}

\begin{figure*}[h!]
    \centering
    \begin{subfigure}[t]{0.47\linewidth}
        \centering
        \begin{tabular}{c}
             \includegraphics[width=\linewidth]{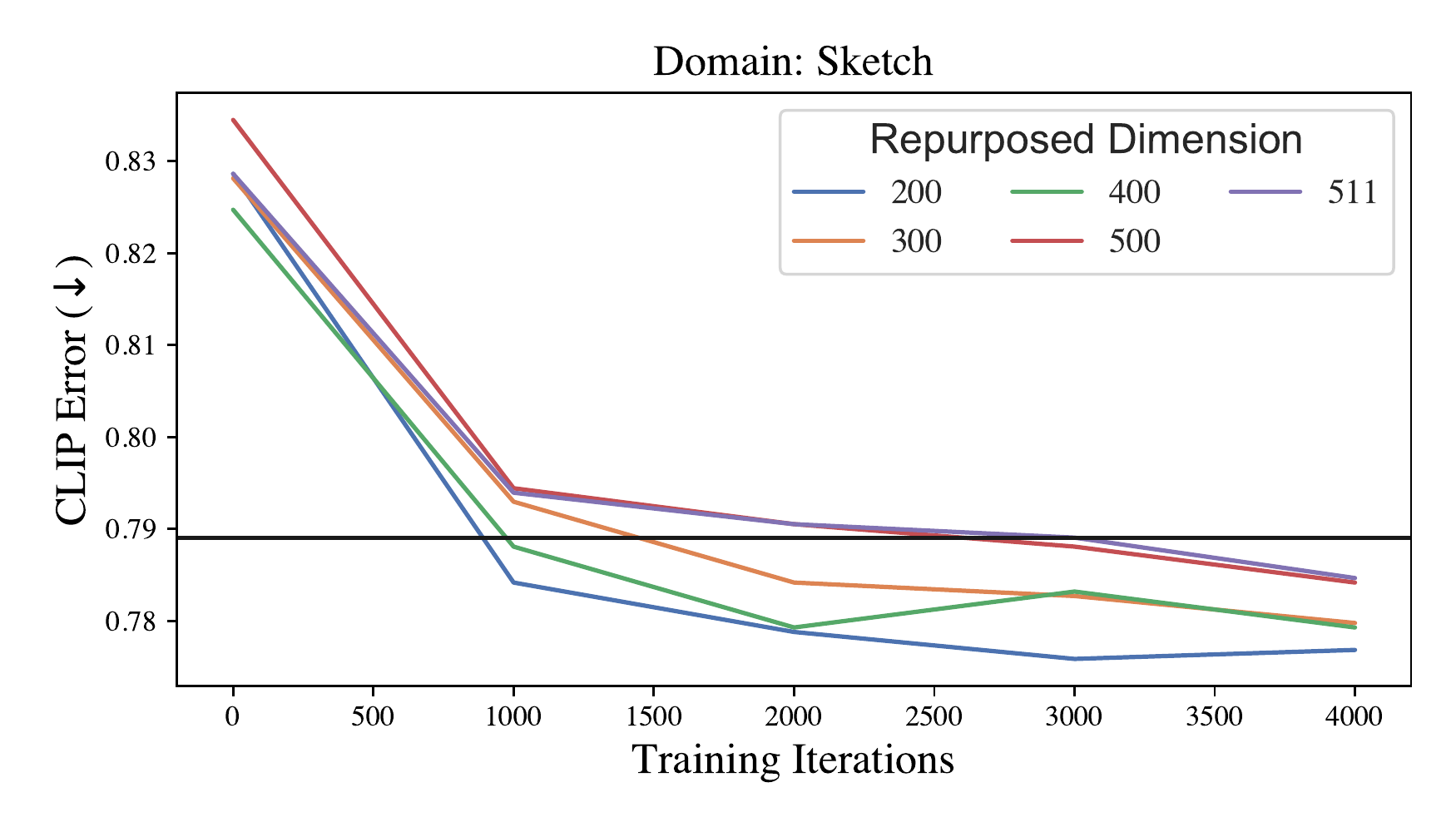}  \\
             \includegraphics[width=\linewidth]{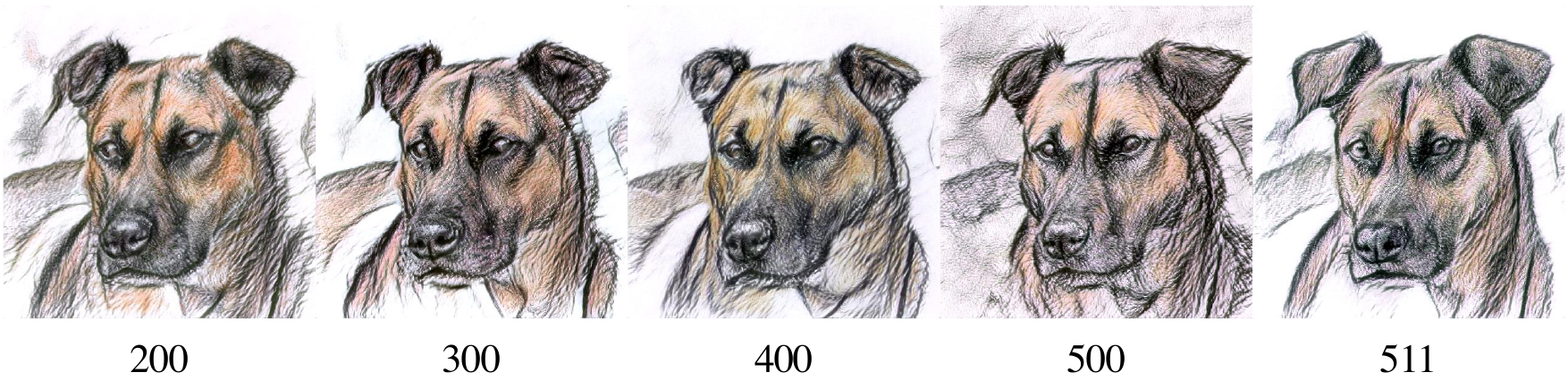}
        \end{tabular}
        \caption{Seed 0}
        \label{fig:dim_quant_sketch}
    \end{subfigure}%
    \;
    \begin{subfigure}[t]{0.47\linewidth}
        \centering
        \begin{tabular}{c}
              \includegraphics[width=\linewidth]{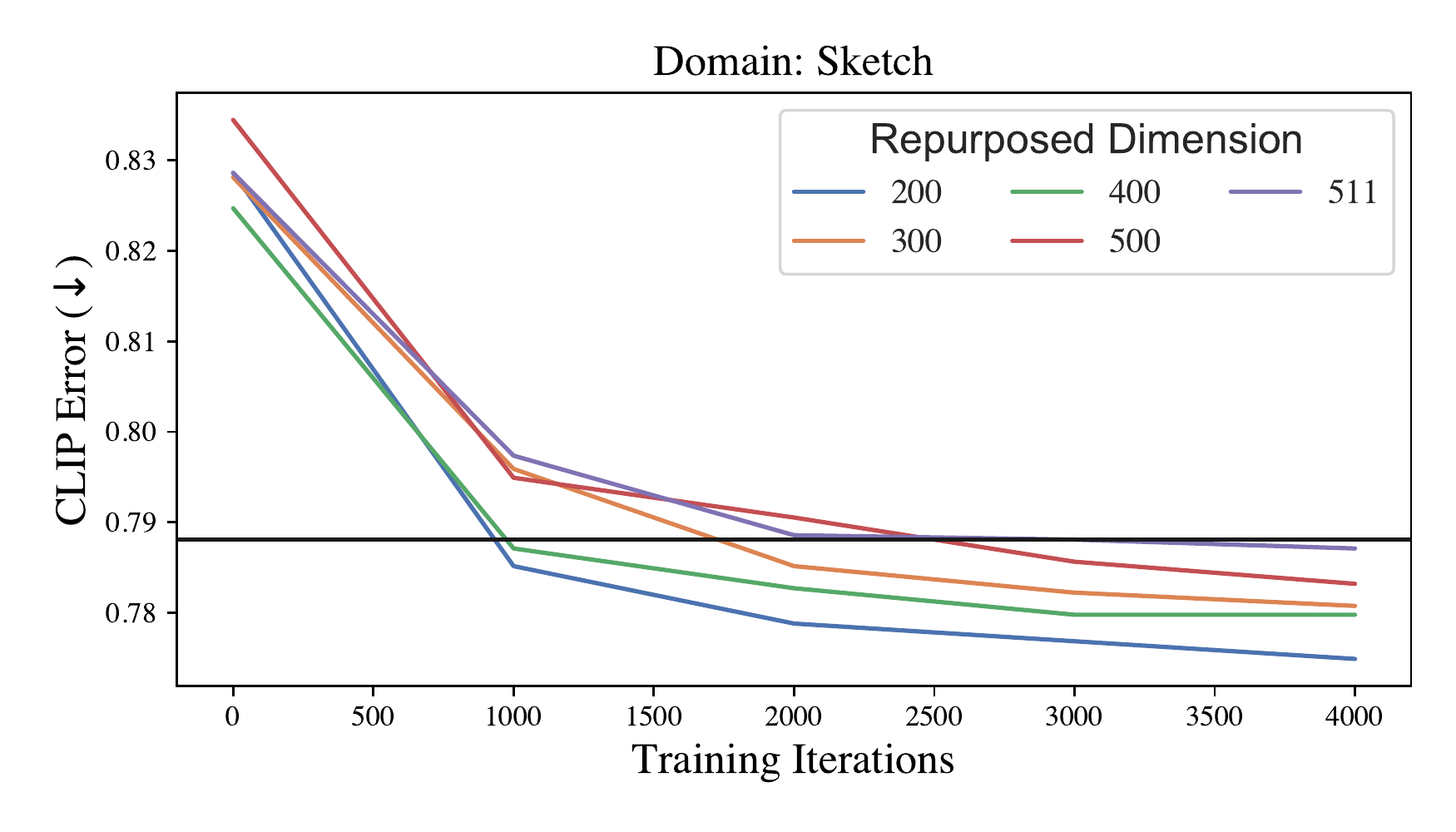} \\
             \includegraphics[width=\linewidth]{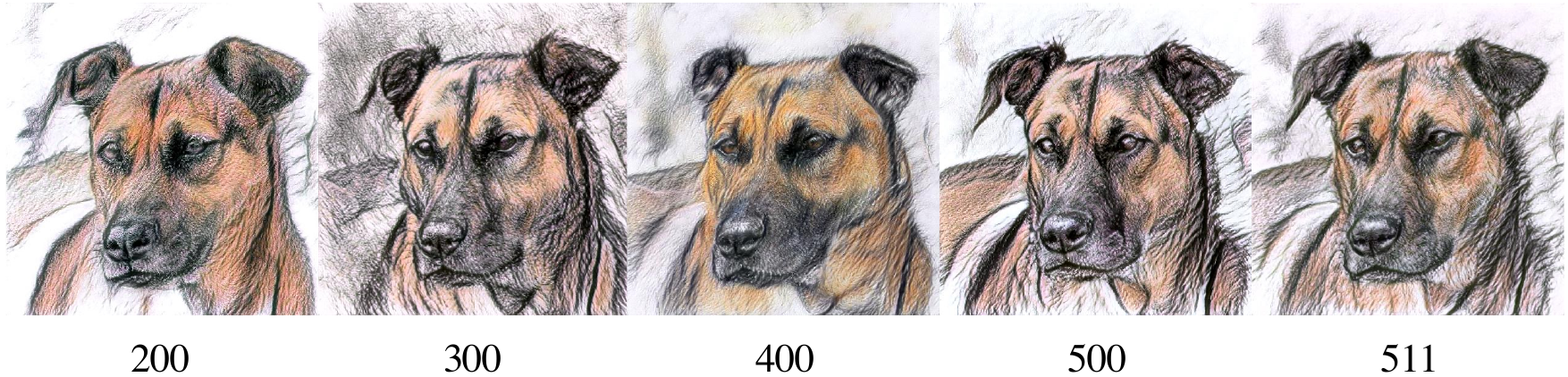}
        \end{tabular}
        \caption{Seed 1}
        \label{fig:dim_qual_sketch}
    \end{subfigure}
    \caption{
    We expand a generator pretrained on AFHQ \cite{choi2020stargan} with 5 domains, varying the dormant direction dedicated to the ``sketch'' domain. We repeat the expansion twice, with different random seeds. Top - reporting CLIP error of images generated from the sketch domain with the text ``a sketch''. Bottom - a sample of generated images from checkpoints obtaining CLIP error closest to the horizontal black line. As can be seen, images generated using different repurposed dimensions differ only slightly. Specifically, changing the random seed induces similar difference.
    }
    \label{fig:effect_dim_sketch}
\end{figure*}

\begin{figure}
    \centering
    \begin{tabular}{c}
        \includegraphics[width=\linewidth]{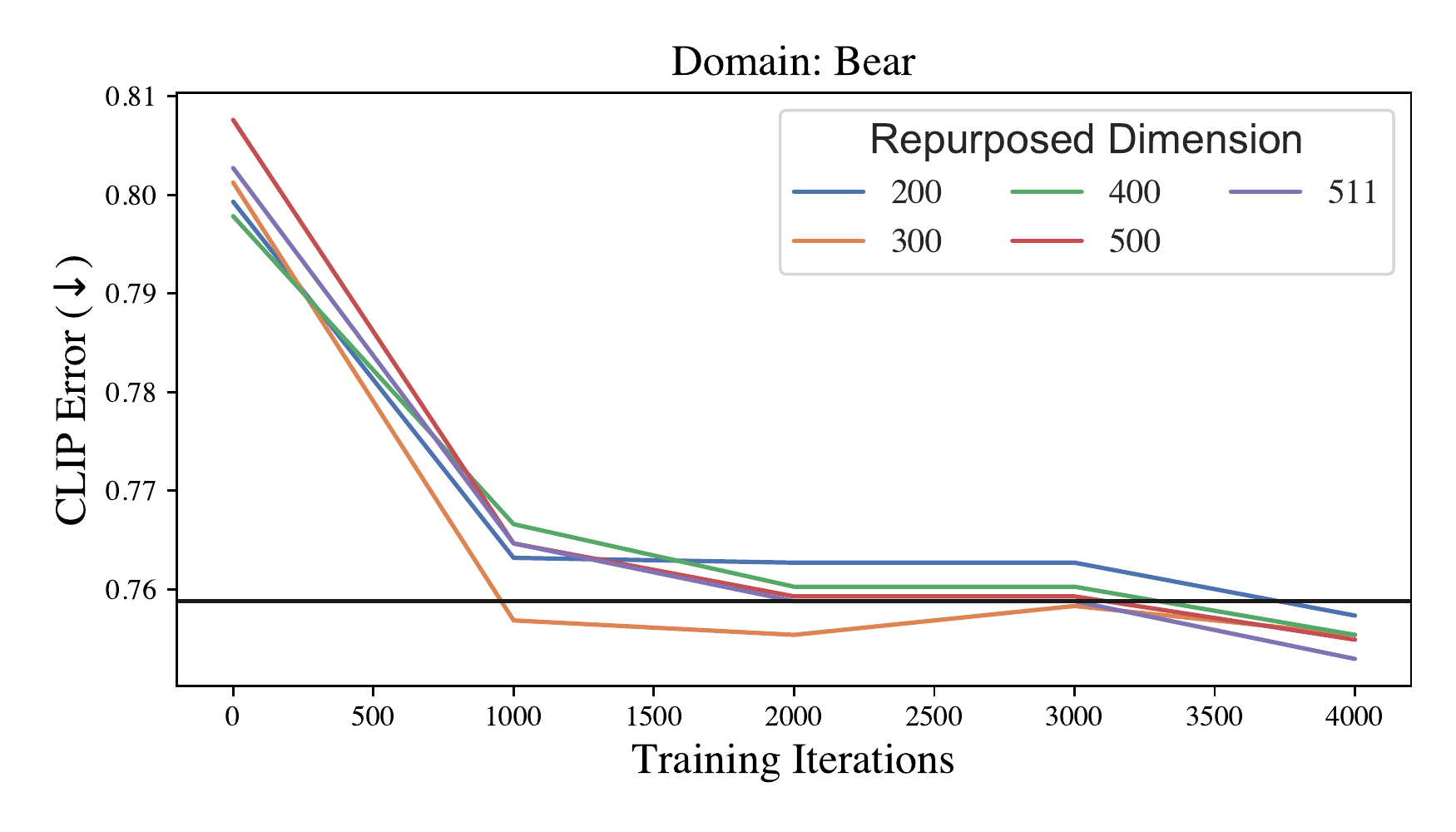}  \\
        \includegraphics[width=\linewidth]{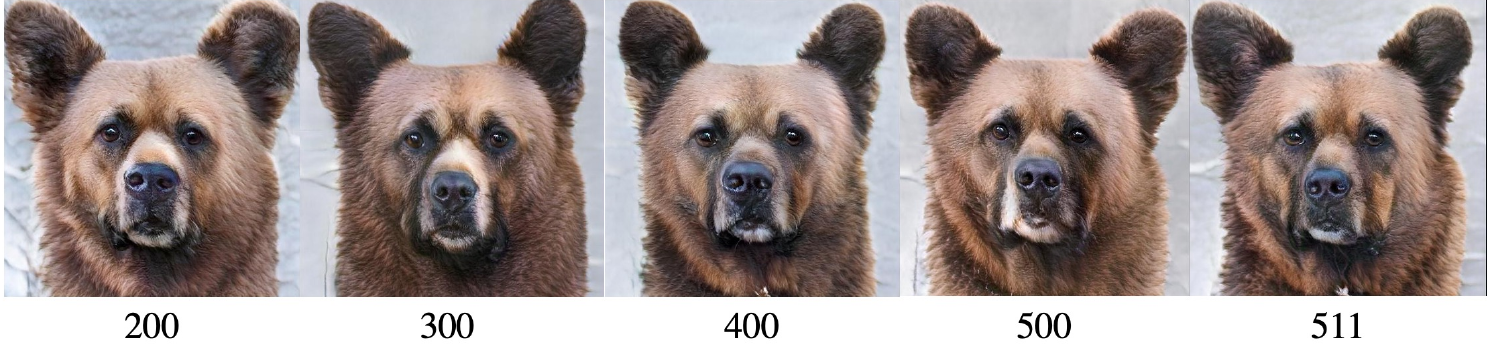}
    \end{tabular}
    \caption{Similar to \cref{fig:effect_dim_sketch}, using a ``bear'' domain instead of ``sketch''. As can be seen, dimensions are ordered differently in terms of minimizing CLIP error, as compared to their order for sketch.}
    \label{fig:effect_dim_bear}
\end{figure}

Our method dedicates a single dormant direction for every newly introduced domain. 
As mentioned in \cref{subsec:setting}, all previous experiments used the \textit{last} dormant directions, sorted in decreasing order according to their corresponding singular values. One might wonder: \textit{Why should one use the last directions?  And among the last directions, how should one match a direction to a domain?}

We now demonstrate that the specific choice of a latent direction has no significant impact on results, as long as it is dormant. 
To this end, we perform multiple expansions, each with 5 new domains introduced by StyleGAN-NADA \cite{gal2021stylegan}, starting from a single generator pretrained on AFHQ \cite{choi2020stargan}. 
For 4 of the new domains -- ``Siberian Husky'', ``Pixar'', ``Funny Dog'', ``Boar'' -- we dedicate the same directions in all experiments. Specifically, we use directions $507-510$, respectively. Directions numbers refer to their location in the decreasingly sorted right-singular vector set. Recall that the dimension of the latent space is $512$, hence these directions are among the last ones.
For the last domain, ``Sketch'', we vary the dedicated direction, using one of the directions $200, 300, 400, 500, 511$. We run the expansion twice with different random seeds.

We study how the choice of direction for the Sketch domain affects its performance. In \cref{fig:effect_dim_sketch} (top) we report the CLIP error of images generated from the ``Sketch'' subspace with the prompt ``a sketch'' as a function of training iterations.
We additionally display sample of generated images from each model in \cref{fig:effect_dim_sketch} (bottom).
As can be seen, similar results are produced from different repurposed directions. Specifically, visual differences observed using different directions, are similar to those observed using the same directions but with different random seeds. This indicates that the differences between directions are negligible and might be entirely due to random chance.

Nevertheless, we do observe that certain directions minimize the CLIP error slightly more efficiently, across random seeds. We therefore run additional 5 expansions, using ``Bear'' instead of ``Sketch''. We now observe a different ordering of directions. We therefore conclude, that even if slight, imperceptible, differences exist between directions, they are not consistent across domains.

In summary, the choice of dormant direction has little to no effect. This result is arguably intuitive, as all dormant directions might be considered equivalent, having insignificant effect on generated images.
Therefore, our choice of using the last directions is almost arbitrary, only motivated by the fact that they are the ``most dormant''. Similarly, no technique is required to match an direction to a domain, and one can simply pick a dormant direction randomly.

\subsection{Repurposing Non-Dormant Directions}
\begin{figure}
    \centering
    \begin{subfigure}{\linewidth}
        \centering
        \begin{tabular}{c}
             \includegraphics[width=\linewidth]{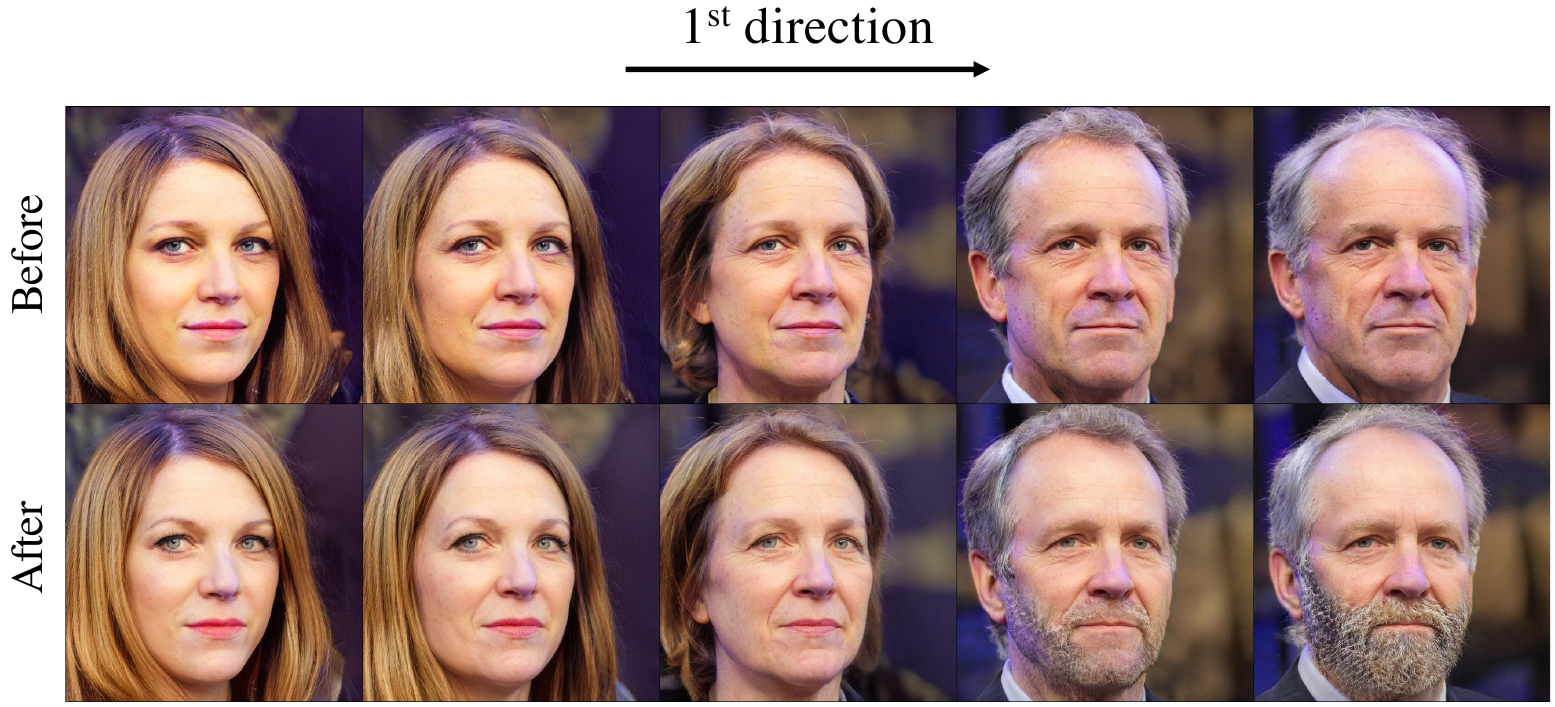} \\
             \includegraphics[width=\linewidth]{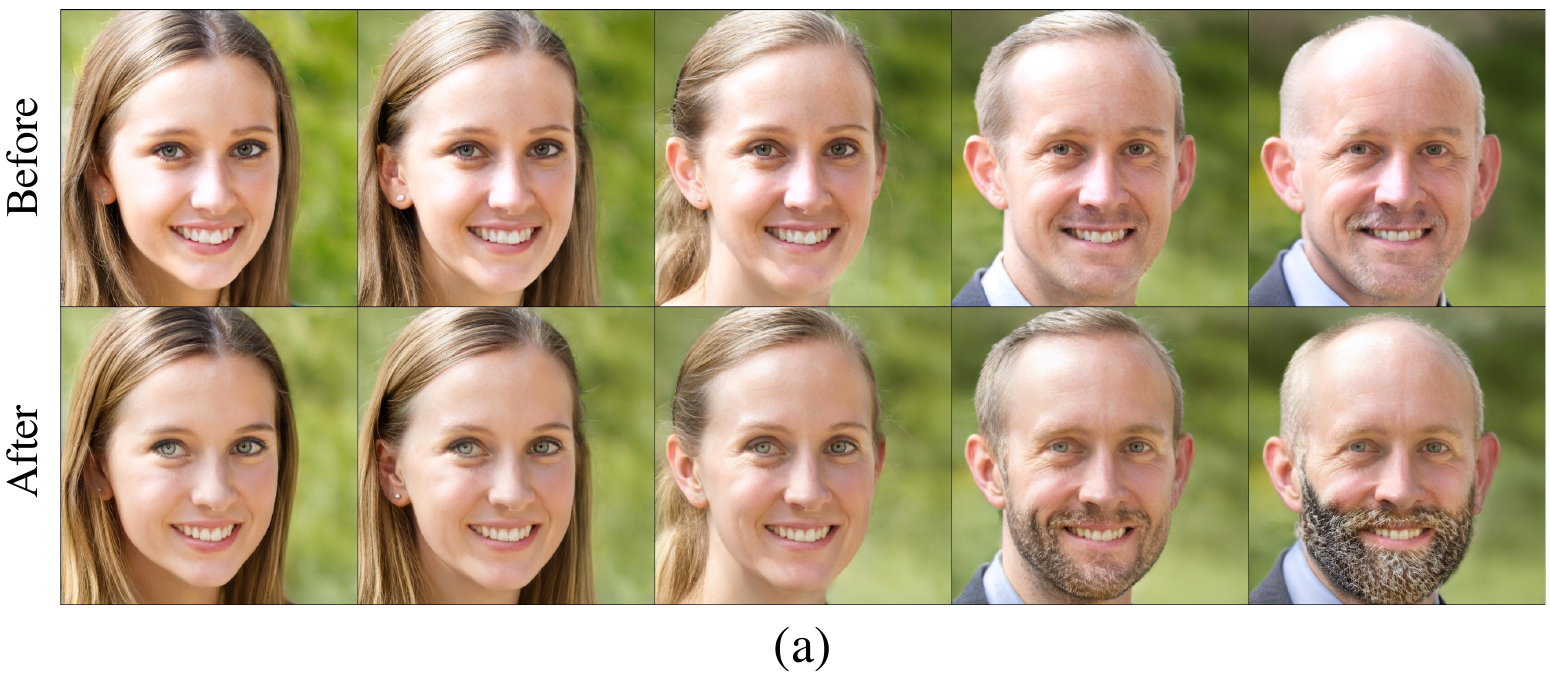}
        \end{tabular}
        \phantomcaption{}
        \label{fig:active_axis_1}
    \end{subfigure}
    
    \vspace{1mm}

    \begin{subfigure}{\linewidth}
        \centering
        \begin{tabular}{c}
             \includegraphics[width=\linewidth]{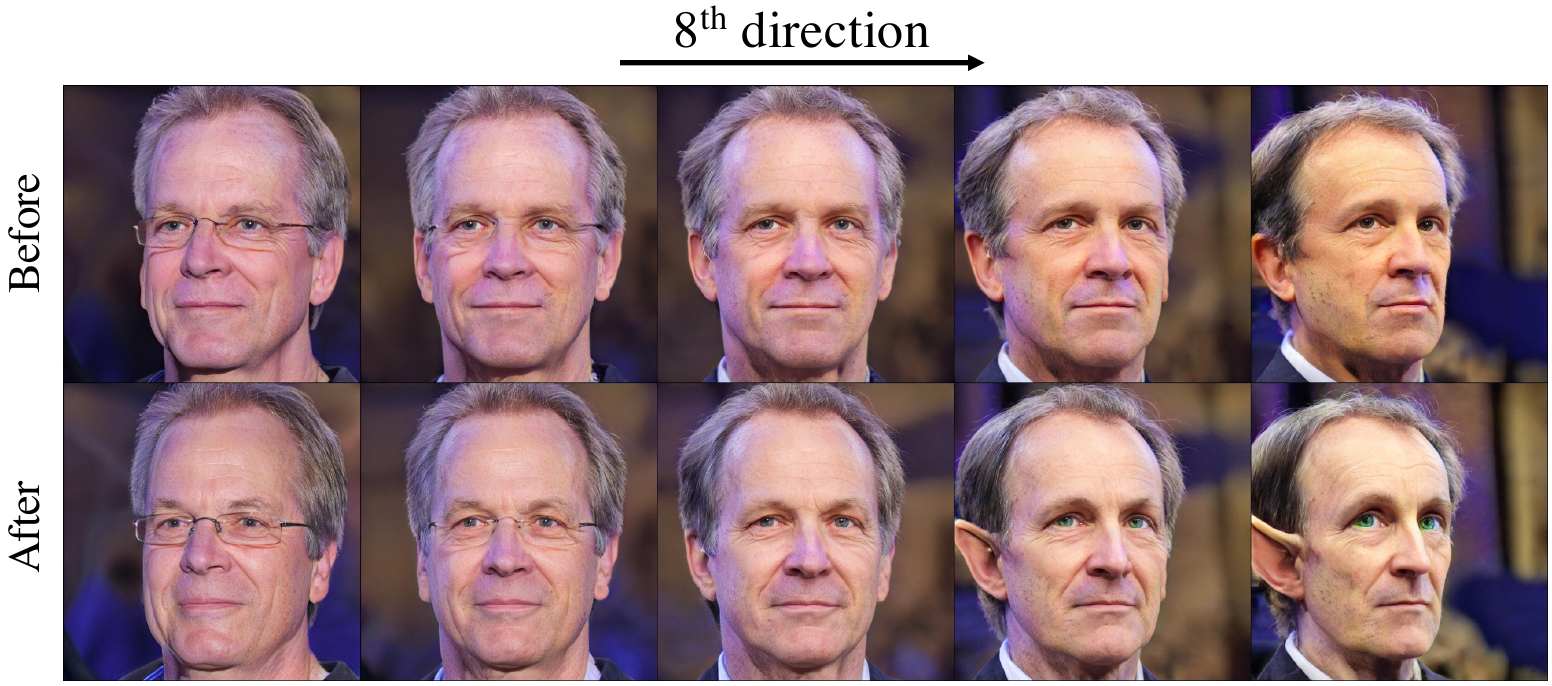} \\
             \includegraphics[width=\linewidth]{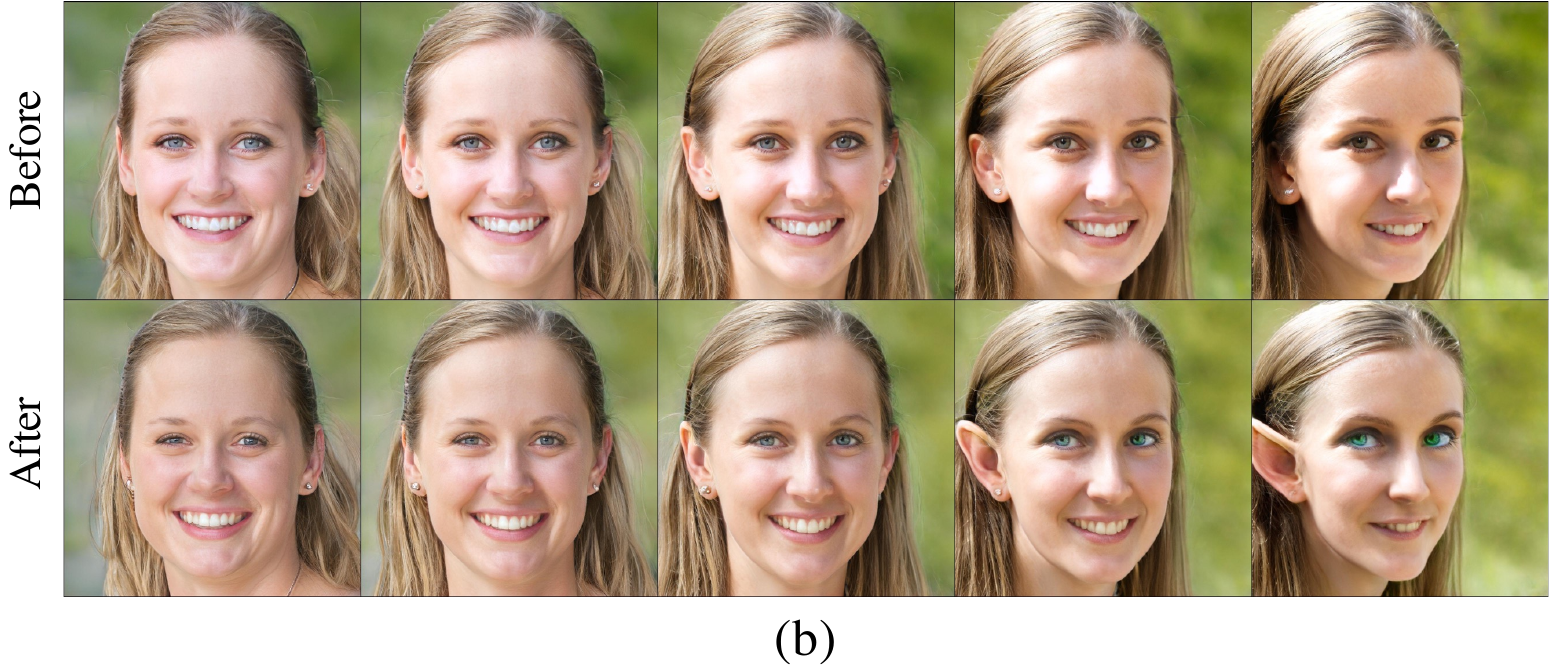}
        \end{tabular}
        \phantomcaption{}
        \label{fig:active_axis_8}
    \end{subfigure}

    \caption{
    Using our training method with non-dormant direction rewrites existing semantic rules and adds new concepts on top of existing ones. (a) Traversing the $1^{st}$ direction originally made people older and more masculine. After fine-tuning, it also adds a beard. (b) Traversing the $8^{th}$ direction originally turned people heads. After fine-tuning it also turns them to elves.
    }
    \label{fig:active}
\end{figure}

Aiming at domain expansion, preserving the source domain is integral. 
Since the non-dormant directions span the variations of the source domains, we explicitly kept them intact, and repurposed only dormant directions.
Nevertheless, the training method itself could be identically applied to non-dormant directions. 
One simply needs to dedicate a non-dormant direction to capture the new domain.
We next demonstrate that applying our method to non-dormant directions is still effective and enables capabilities beyond domain expansion.

Traversing the $1^{st}$ latent direction in the generator pretrained on FFHQ \cite{karras2019style}, makes people in generated images appear older and more masculine. Some users might decide that they associate having a full beard with being older and more masculine. To support such behavior, we fine-tune the generator with a transformed StyleGAN-NADA \cite{gal2021stylegan}, to capture ``a person with a beard'' along the $1^{st}$ direction. 
We display images generated along traversals of the $1^{st}$ direction, before and after tuning, in \cref{fig:active_axis_1}. As can be seen, the generator now represents having a beard, along its $1^{st}$ latent direction, in addition to its previous behavior. 

The capability to add new concepts in addition to existing ones does not depend on the close relationship between the two in the last examples. To demonstrate this point, we tune the generator to capture ``Elf'' along its $8^{th}$ direction, which originally encodes head pose (and a few other properties). Results are displayed in \cref{fig:active_axis_8}.

Previous results are clearly not solving domain expansion, as they alter the original behavior of the source domain.
Instead, one might say they adapt the domain modeled by the generator. Nevertheless, there exists a profound difference to existing domain adaptation methods. Our resulting generator does not completely overriding the source domain. Instead, in a precise and controllable manner, it modifies individual factors of variation. 
Therefore, a user can carefully \textit{rewrite} \cite{wang2022rewriting,bau2020rewriting} the semantic rules of a generative model, allowing greater control and freedom. 

\subsection{Distance to Repurposed Subspace}
\label{subsec:effect_of_s}
\begin{figure*}
    \centering
    \begin{subfigure}{\linewidth}
         \begin{tabular}{cc}
              \includegraphics[width=0.48\linewidth]{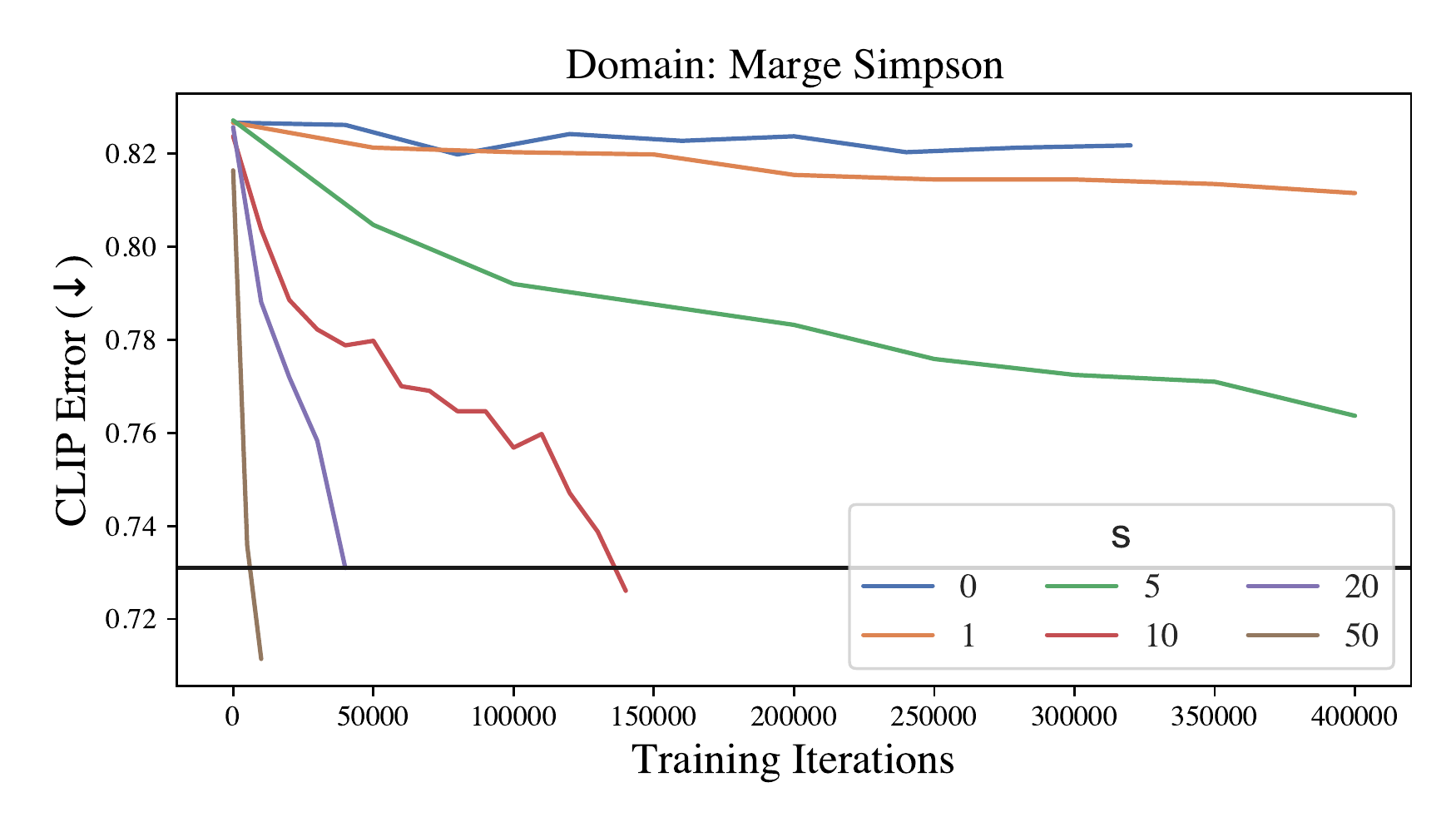} &
              \includegraphics[width=0.48\linewidth]{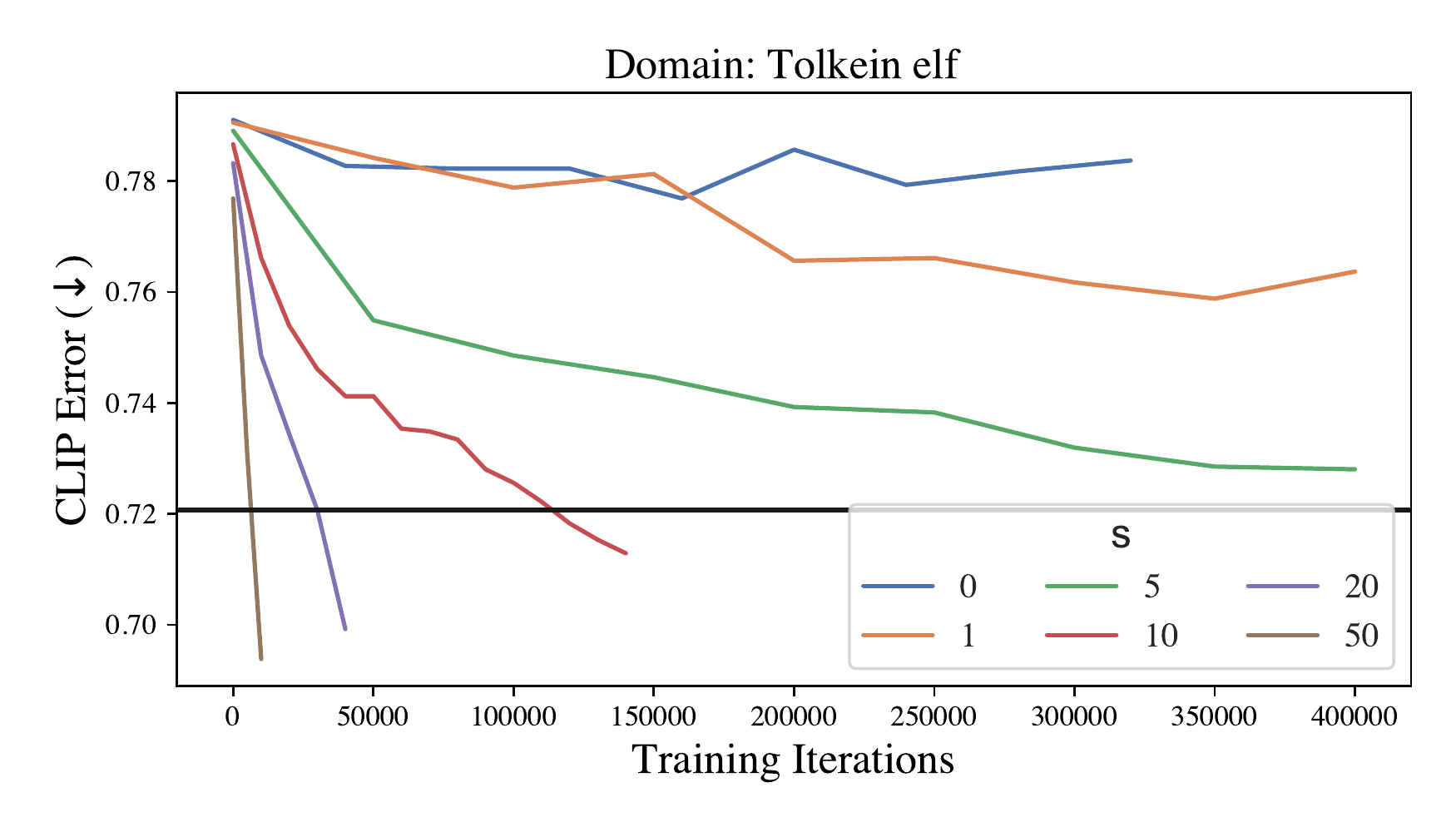}
         \end{tabular}
         \vspace{-2mm}
         \caption{}
         \label{fig:effect_of_s_quant}
    \end{subfigure}
    \vspace{1mm}
    \begin{subfigure}{\linewidth}
        \centering
        \vspace{2mm}
        \includegraphics[width=0.95\linewidth]{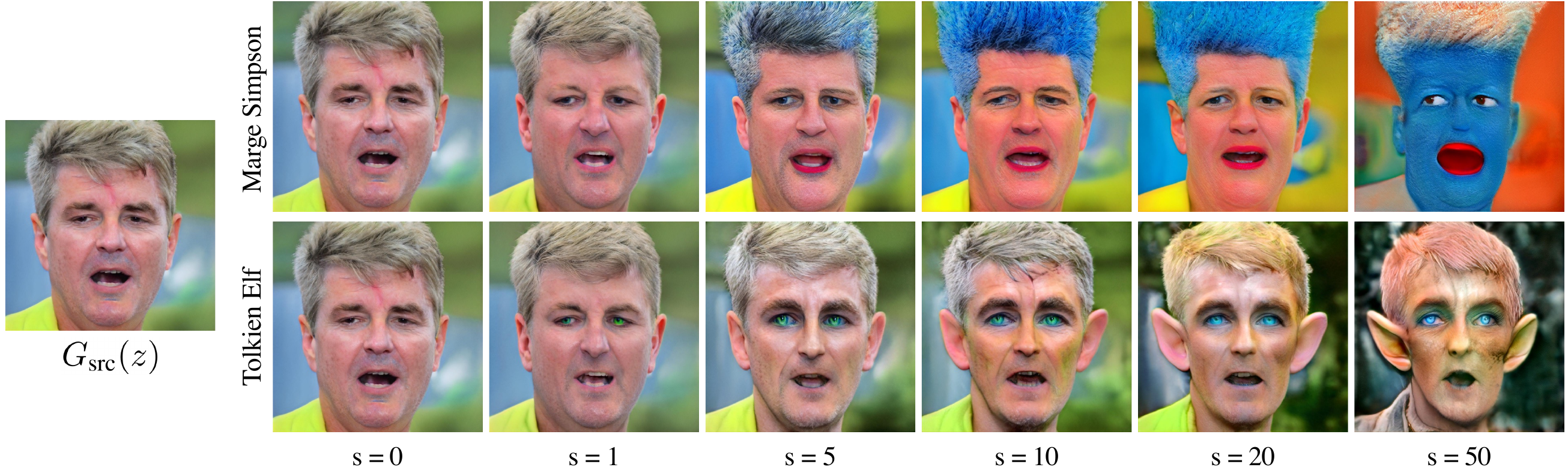}
        \caption{}
        \label{fig:effect_of_s_qual}
    \end{subfigure}
    \caption{Evaluating the effect of the distance between the base and repurposed subspace, $s$. (a) We compare CLIP error as a function of training iterations, between models trained with different values of parameter $s$. (b) Generated images from models having CLIP error as close as possible to the black horizontal line. As can be seen, increasing $s$ corresponds to faster minimization of CLIP error. However, even with comparable CLIP errors, visual effect might vary significantly. Large values of parameter $s$ are often associated with undesired artifacts. We find that values between $[10,30]$ are usually preferable.}
    \label{fig:effect_of_s}
\end{figure*}

\begin{figure}
    \centering
    \includegraphics[width=\linewidth]{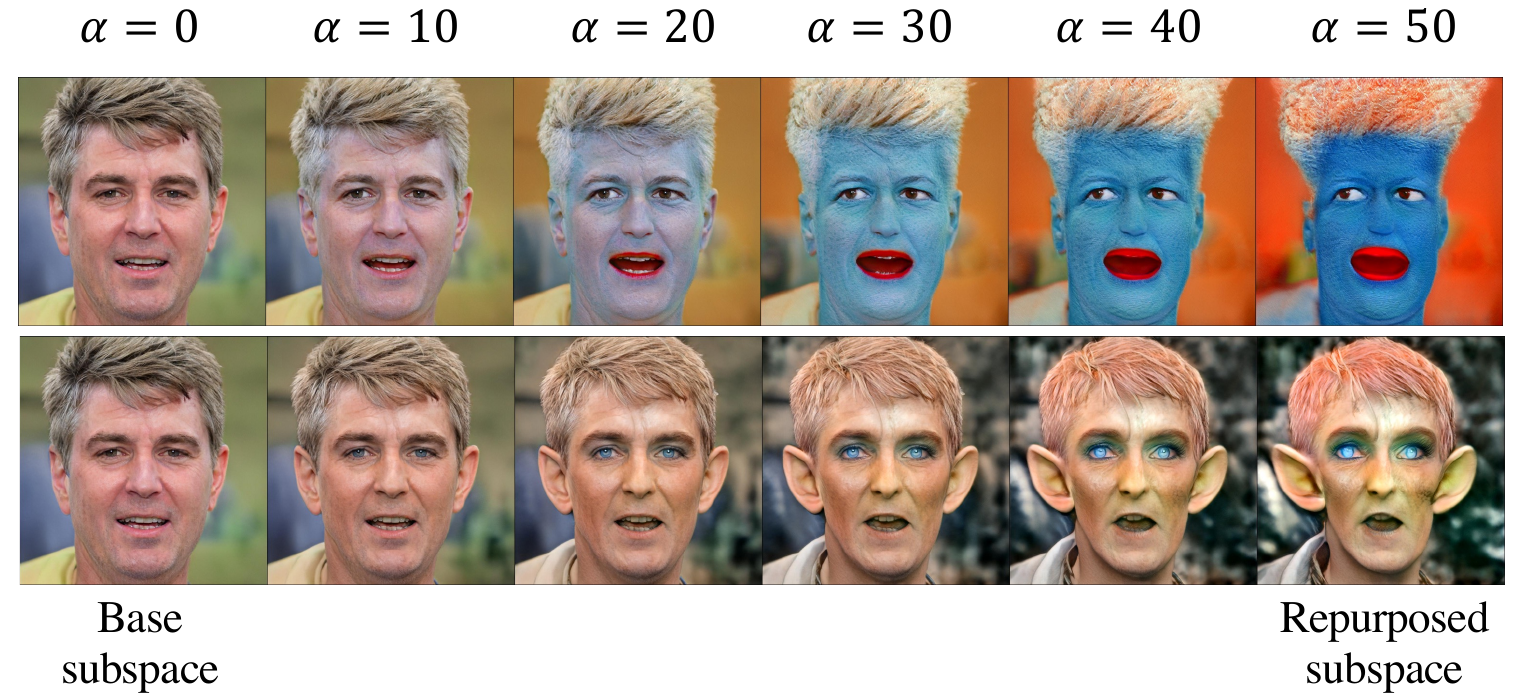}
    \caption{Interpolation between the base subspace and the repurposed subspace where $s=50$. As can be seen, undesired behavior occurring at repurposed subspace (\eg blue skin Marge Simpson) cannot be mitigated by traversing shorter distances in test time. The choice of parameter $s$ is crucial in training time.}
    \label{fig:effect_of_s_interpolation}
\end{figure}

Repurposed subspaces are defined by transporting the base subspace along a linear direction by a predetermined scalar size $s$ (See \cref{eq:repurposed_space} in the main paper). All results in the paper, across domains and variations used $s=20$.
We next evaluate the effect the hyperparameter $s$ has on results.
To this end, we perform multiple expansions of an FFHQ \cite{karras2019style} generator with $100$ new variations, while varying the value of $s$.

We measure CLIP errors (introduced in \cref{subsec:effect_multi}) of images generated from repurposed subspaces and the corresponding target text used for training, as a function of training iterations. In \cref{fig:effect_of_s_quant} we report the results for two variations - ``Marge Simpson'' and ``Tolkein Elf''.
As can be seen, for all $s>0$, CLIP error decreases as training progresses, and it decreases ``faster'' for greater values of the parameter $s$. Even with $\times 10$ more iterations, the model trained with $s=5$ does not reach the CLIP error of the model trained with $s=20$.

Images generated from the repurposed subspaces are displayed in \cref{fig:effect_of_s_qual}. For each value of $s$, we use the checkpoint that resulted in the closest CLIP error to that obtained by a favored $s=20$ checkpoint.
As can be seen, not only training time is affected by parameter $s$, but the visual effects captured by training vary significantly. 

We observe that models trained with greater values of parameter $s$ depict a more significant change with respect to the source domain. 
When parameter $s$ is too small (\eg, $s \leq 5$), the model captures only few, simple characteristics of the new domain. On the other hand, when parameter $s$ is too large (\eg, $s = 50$), the model commonly generates images that are blurry, have color artifacts or even do not capture the target text well. For example, with the target text ``Marge Simpson'', the model learns to generate images with blue skin rather than blue hair.
We note that these undesired artifacts cannot be mitigated by training with a large value of parameter $s$ originally, and use a smaller one in test-time, as demonstrated in \cref{fig:effect_of_s_interpolation}.

Following these results, we conclude that the parameter $s$ has a regularizing effect. Placing the domains ``closer'' in the latent space causes them to be more similar in image space as well. Conversely, placing the domains further apart allows the new domain to capture more drastic, out-of-domain effects. 

Eventually, choosing a value for parameter $s$ is subject to user preference. In our experiments, we have found that values in the range of $[10,30]$ offer satisfying results, across different source and expanded domains.

We last note that the regularization effect of parameter $s$ could be explained by the existence of a globally consistent ``pace of change'' of the generator with respect to the latent space. With StyleGAN, such behavior is explicitly encouraged using a Perceptual Path Length (PPL) regularization term \cite{karras2020analyzing}. Nevertheless, we observe identical results when omitting this regularization during our expansion. 

\subsection{How Many Domains Can Fit?}
\begin{figure*}
    \centering
    \includegraphics[width=0.9\linewidth]{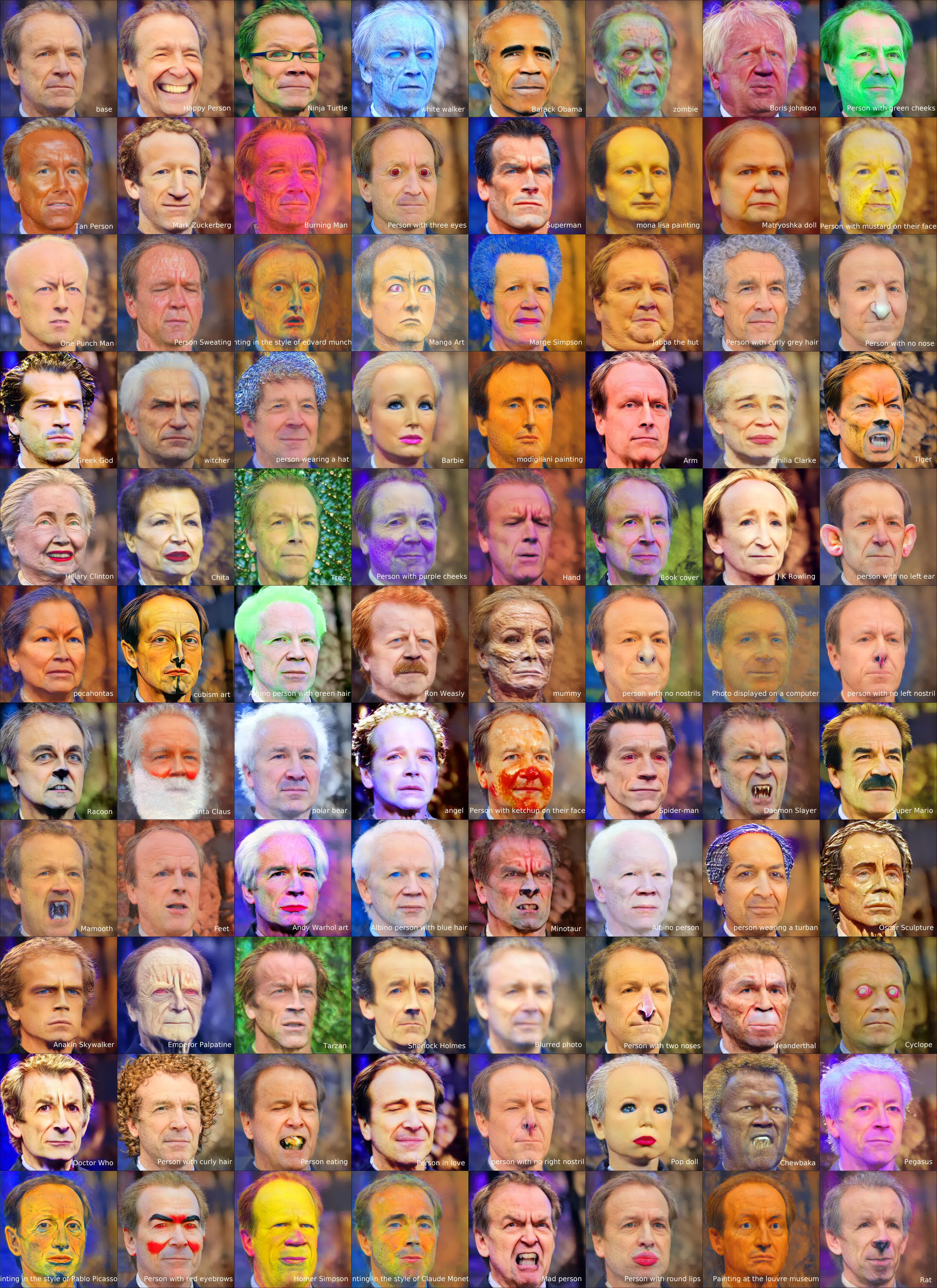}
    \caption{Subset 1/3 of generated images from a model expanded with $400$ domains.}
    \label{fig:many_domains_1}
\end{figure*}

\begin{figure*}
    \centering
    \includegraphics[width=0.9\linewidth]{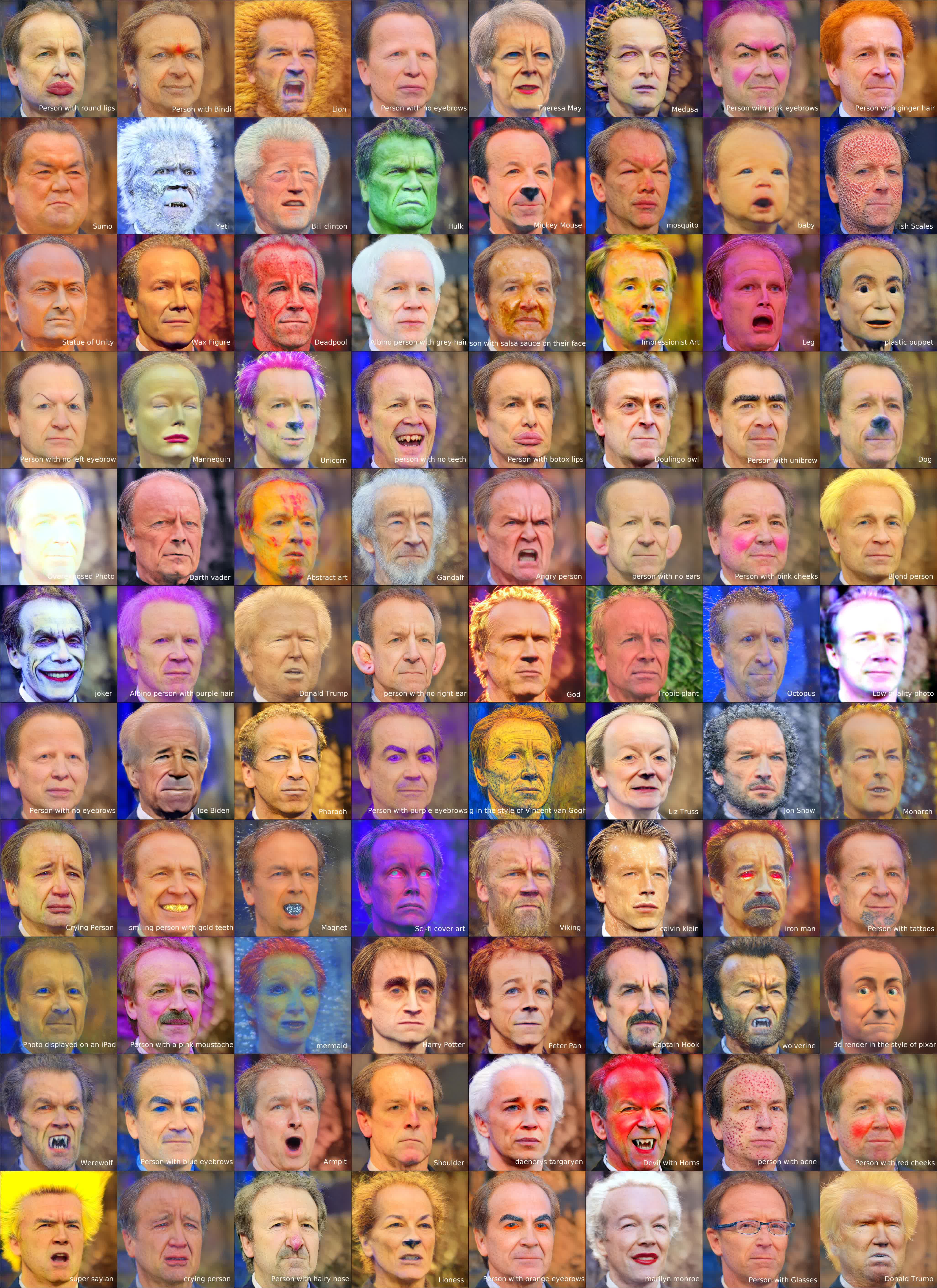}
    \caption{Subset 2/3 of generated images from a model expanded with $400$ domains.}
    \label{fig:many_domains_2}
\end{figure*}

\begin{figure*}
    \centering
    \includegraphics[width=0.9\linewidth]{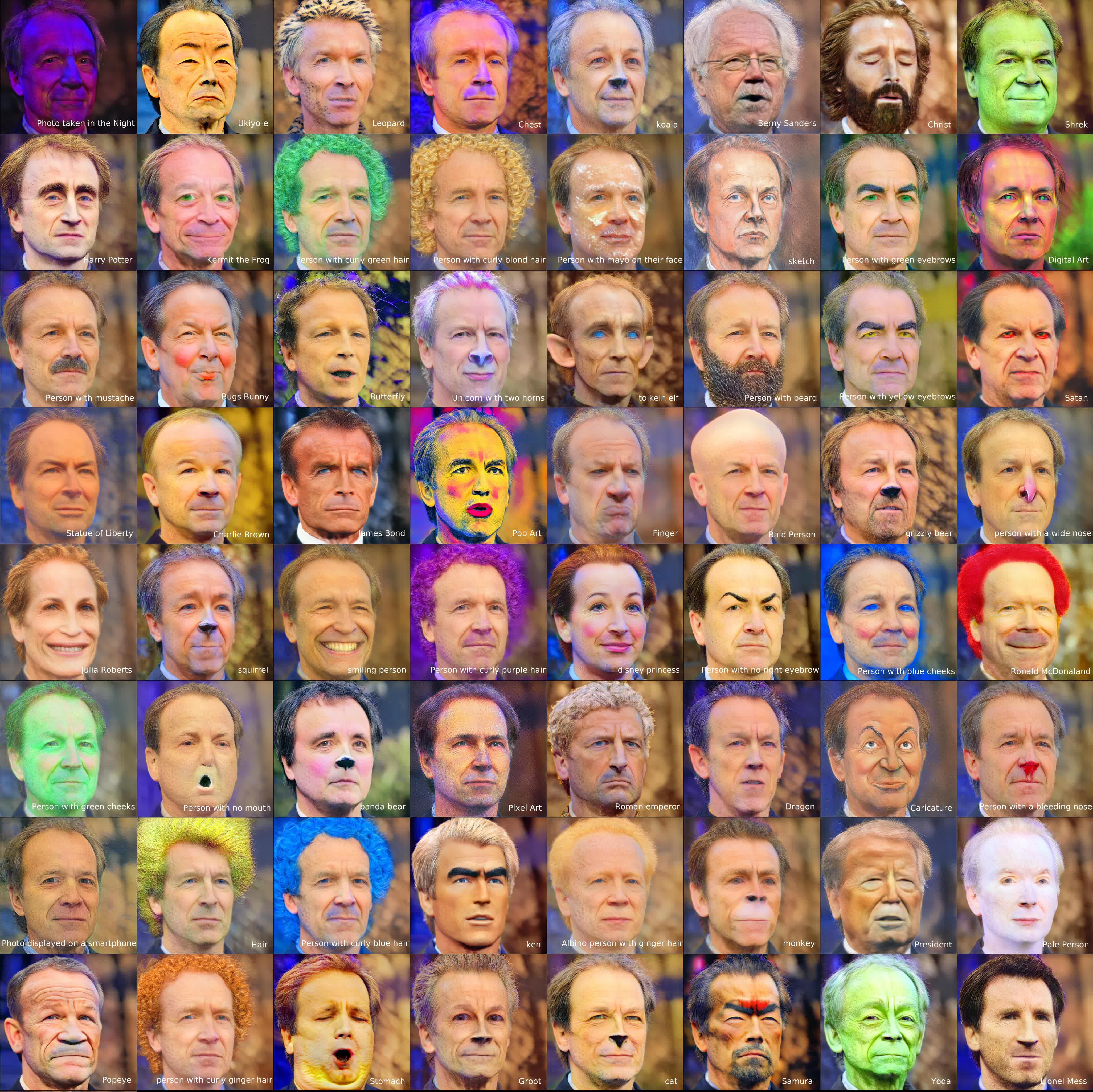}
    \caption{Subset 3/3 of generated images from a model expanded with $400$ domains.}
    \label{fig:many_domains_3}
\end{figure*}

So far, the largest number of new domains used for expansion was $105$. 
The results from \cref{subsec:analysis} indicated that there might be up to $400$ dormant directions. 
\textit{Could they all be repurposed?}

We apply our method to expand a generator pretrained on FFHQ with $400$ new domains, repurposing the last (and perhaps all) dormant directions. Incredibly, the expansion succeeds. 
We find that the expansion follows the same findings discussed in \cref{subsec:effect_multi} -- training is slower, yet quality is uncompromised. Specifically, the FID score from the base subspace is $2.83$ compared to $2.80$ in our model expanded with $105$ domains. We  display images generated from this model in the accompanying video and in \cref{fig:many_domains_1,fig:many_domains_2,fig:many_domains_3}.

\subsection{Additional Compositions Results}
\begin{figure*}
    \centering
    \begin{tabular}{cc}
         \includegraphics[width=0.5\linewidth]{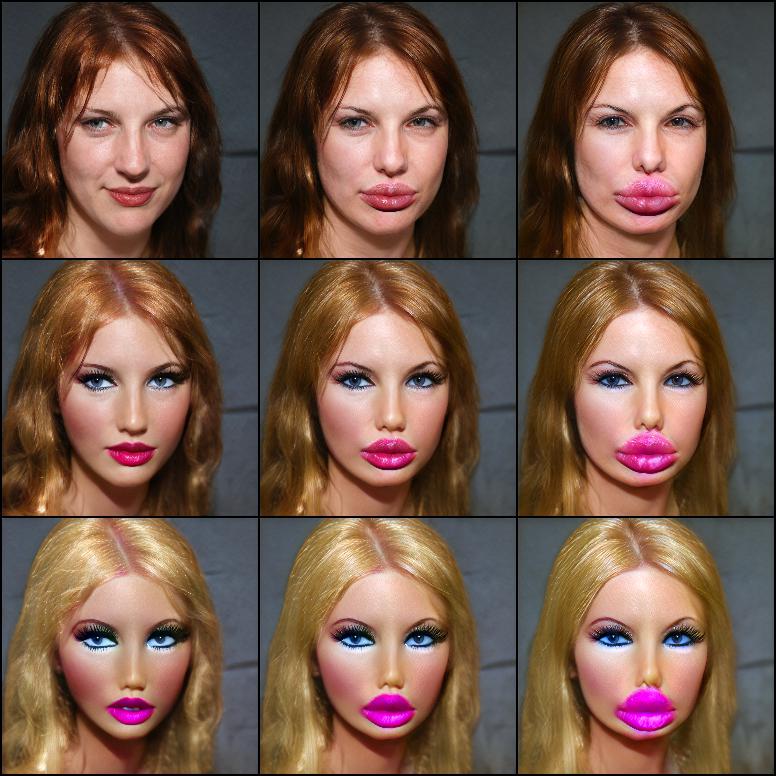} &
         \includegraphics[width=0.5\linewidth]{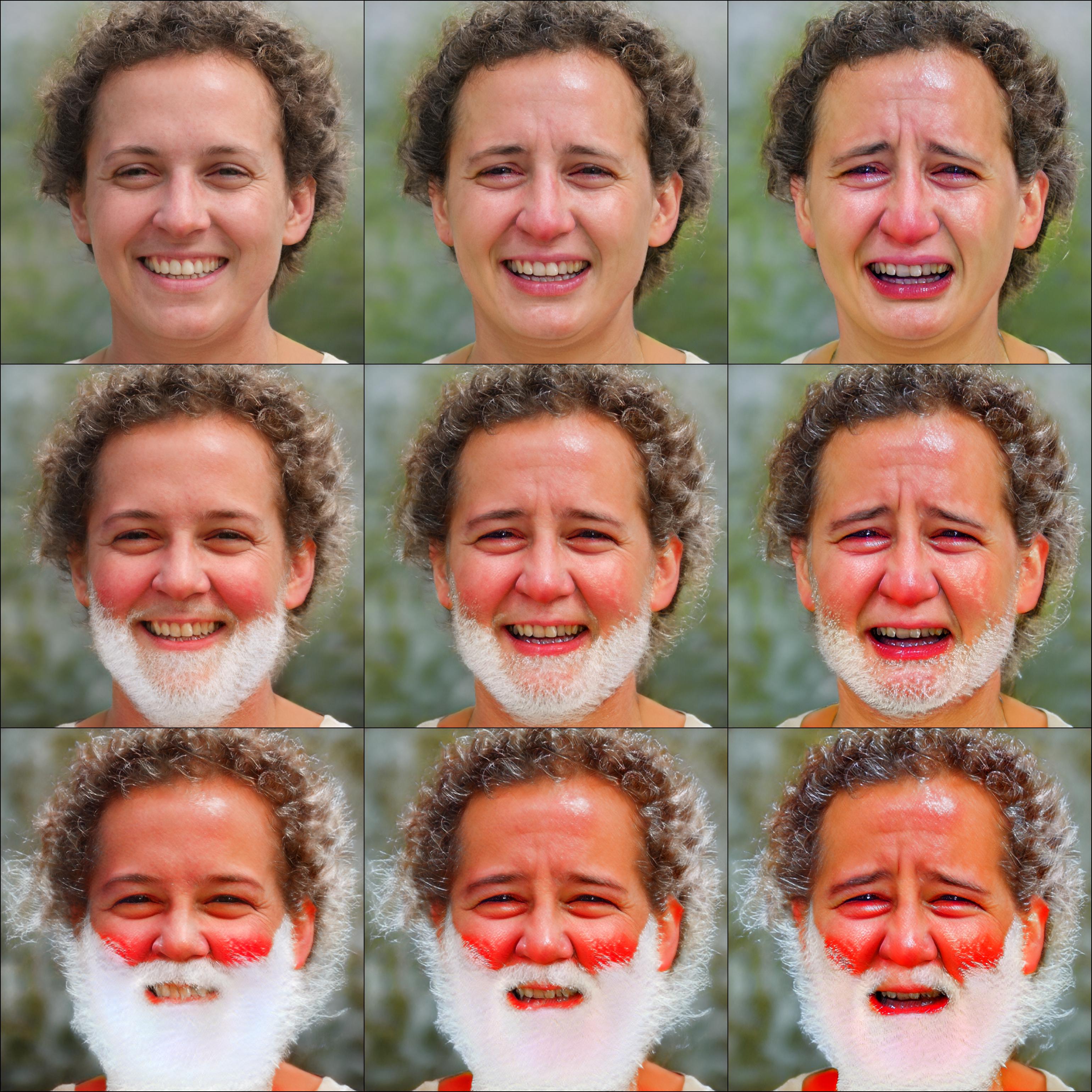}
         \\
         \includegraphics[width=0.5\linewidth]{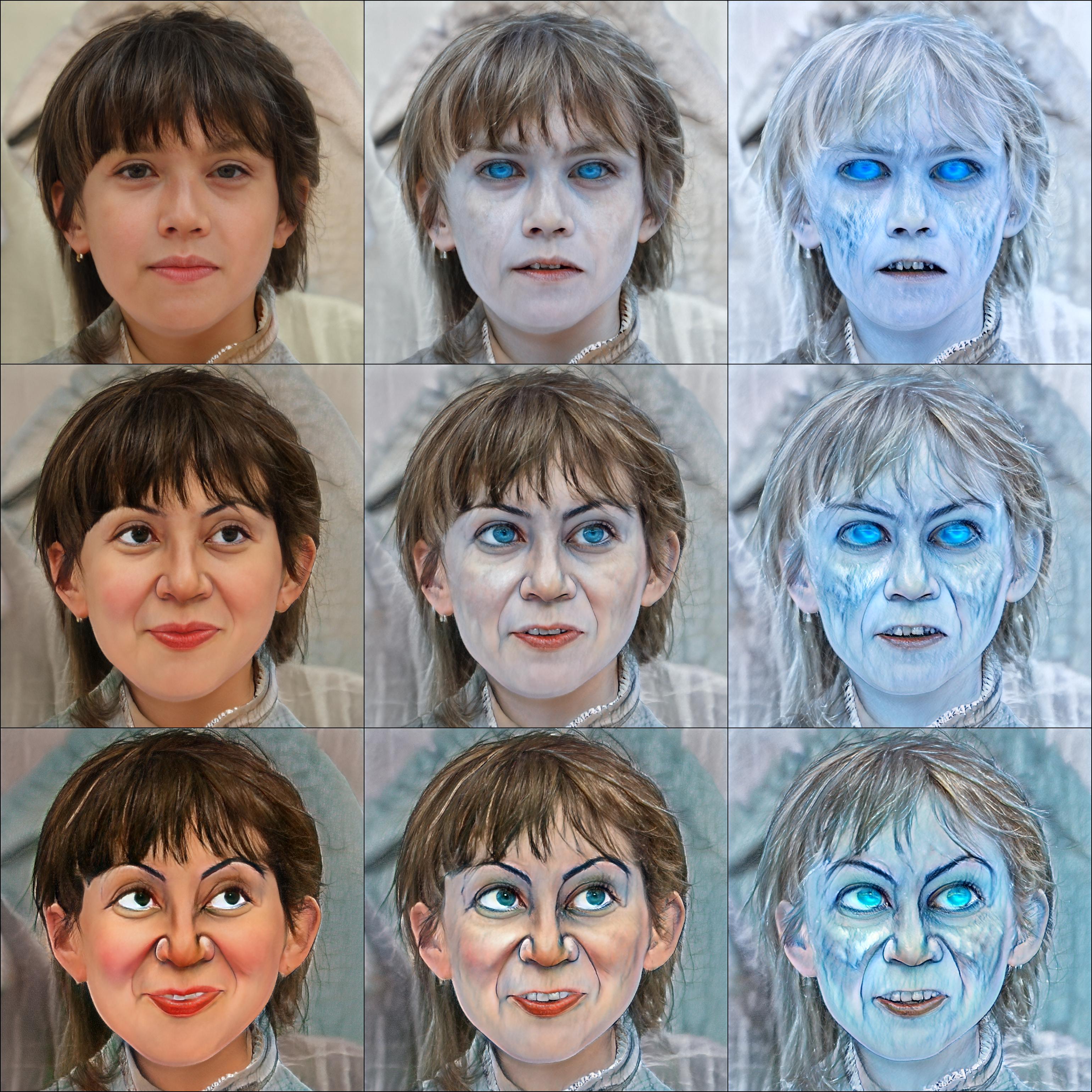} &
         \includegraphics[width=0.5\linewidth]{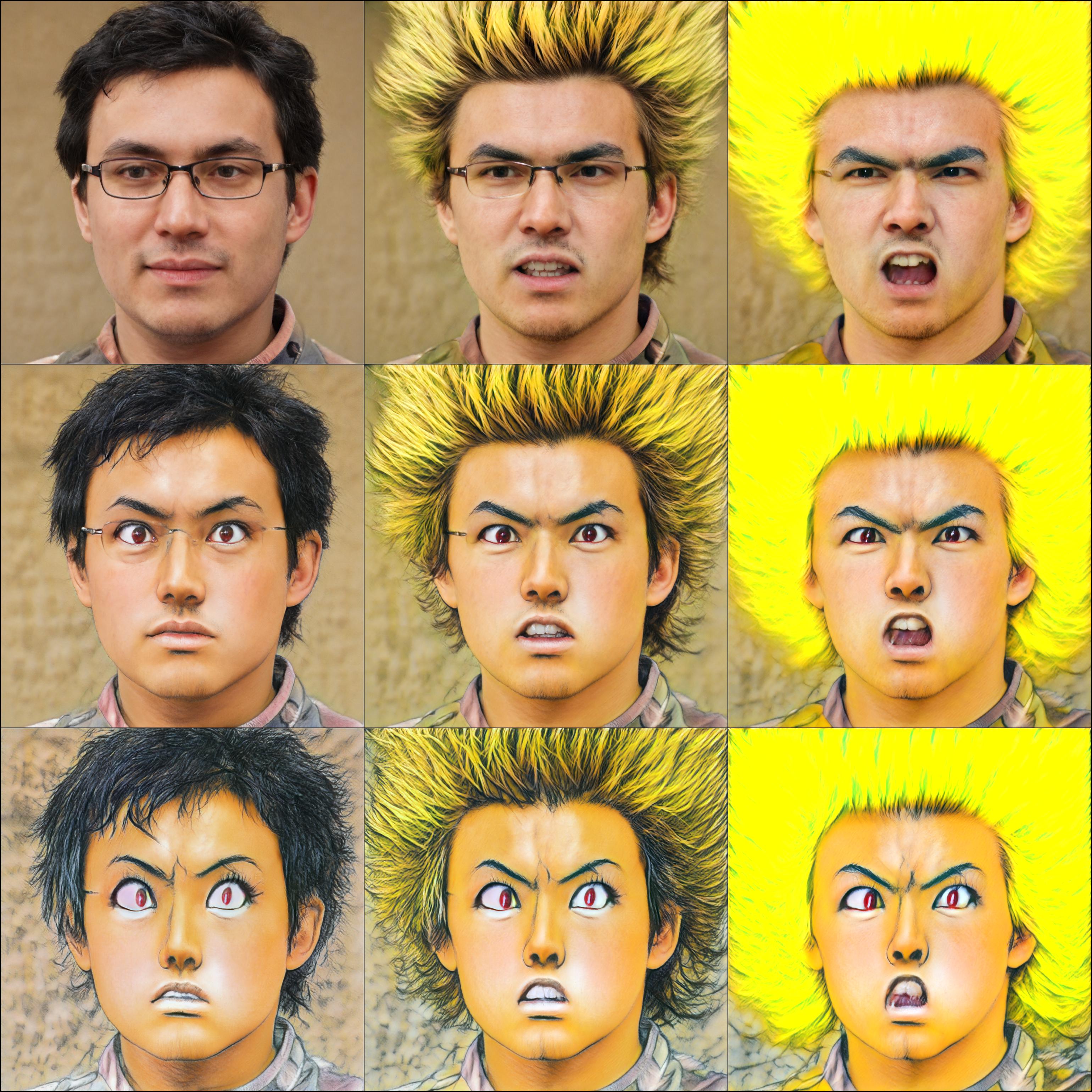}
    \end{tabular}
    \caption{Composition of factors of variation introduced to a generator pretrained on FFHQ \cite{karras2019style}. Following the format of \cref{fig:composition_showcase}}.
    \label{fig:ffhq_composition_supp}
\end{figure*}

\begin{figure*}
    \centering
    \begin{tabular}{cc}
         \includegraphics[width=0.5\linewidth]{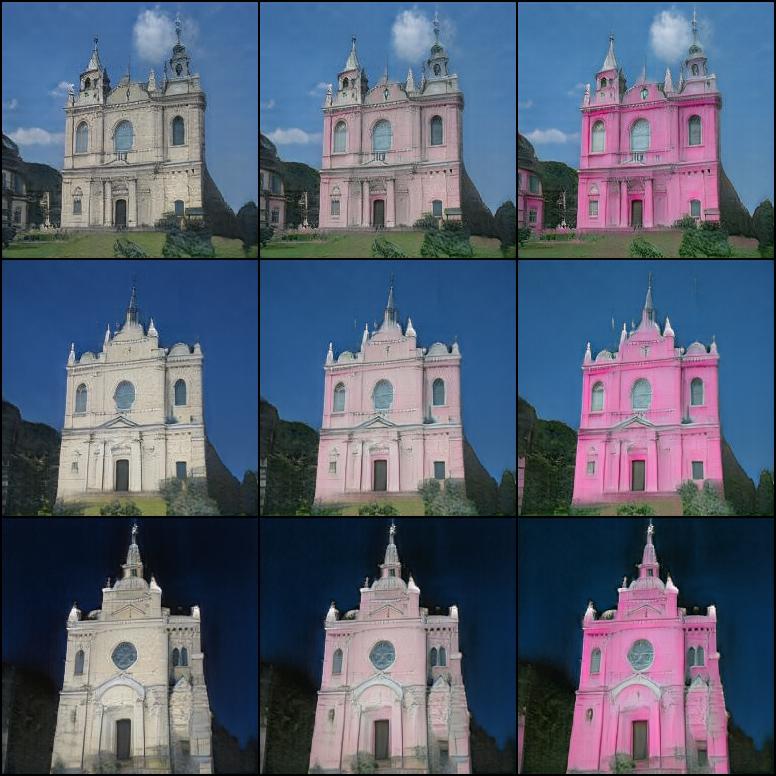} &
         \includegraphics[width=0.5\linewidth]{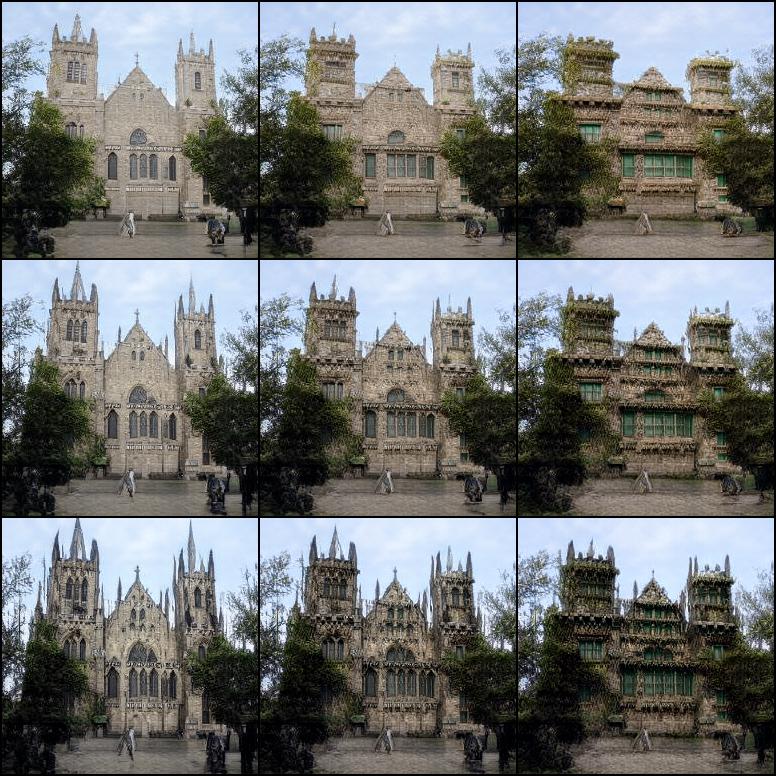}
         \\
         \includegraphics[width=0.5\linewidth]{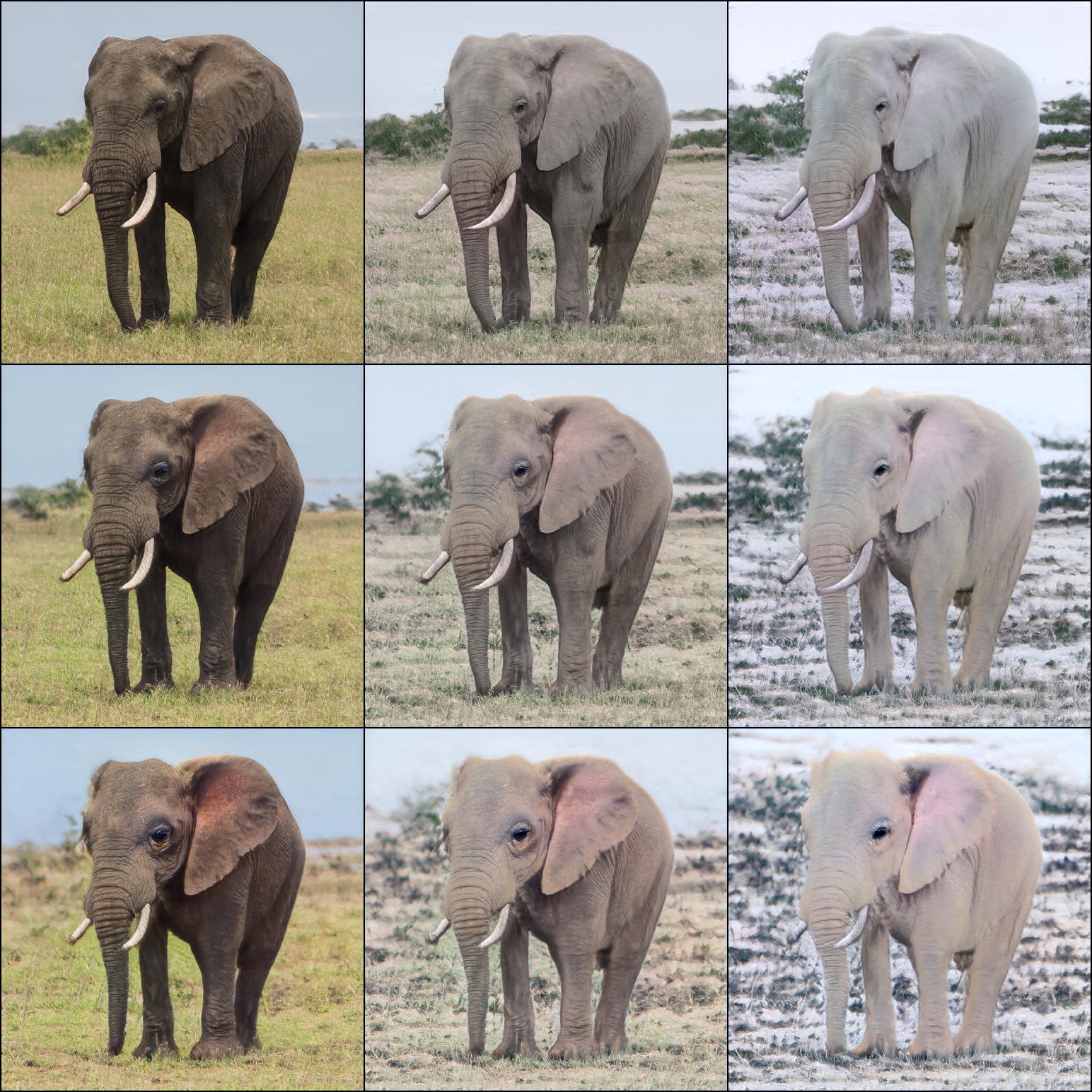} &
         \includegraphics[width=0.5\linewidth]{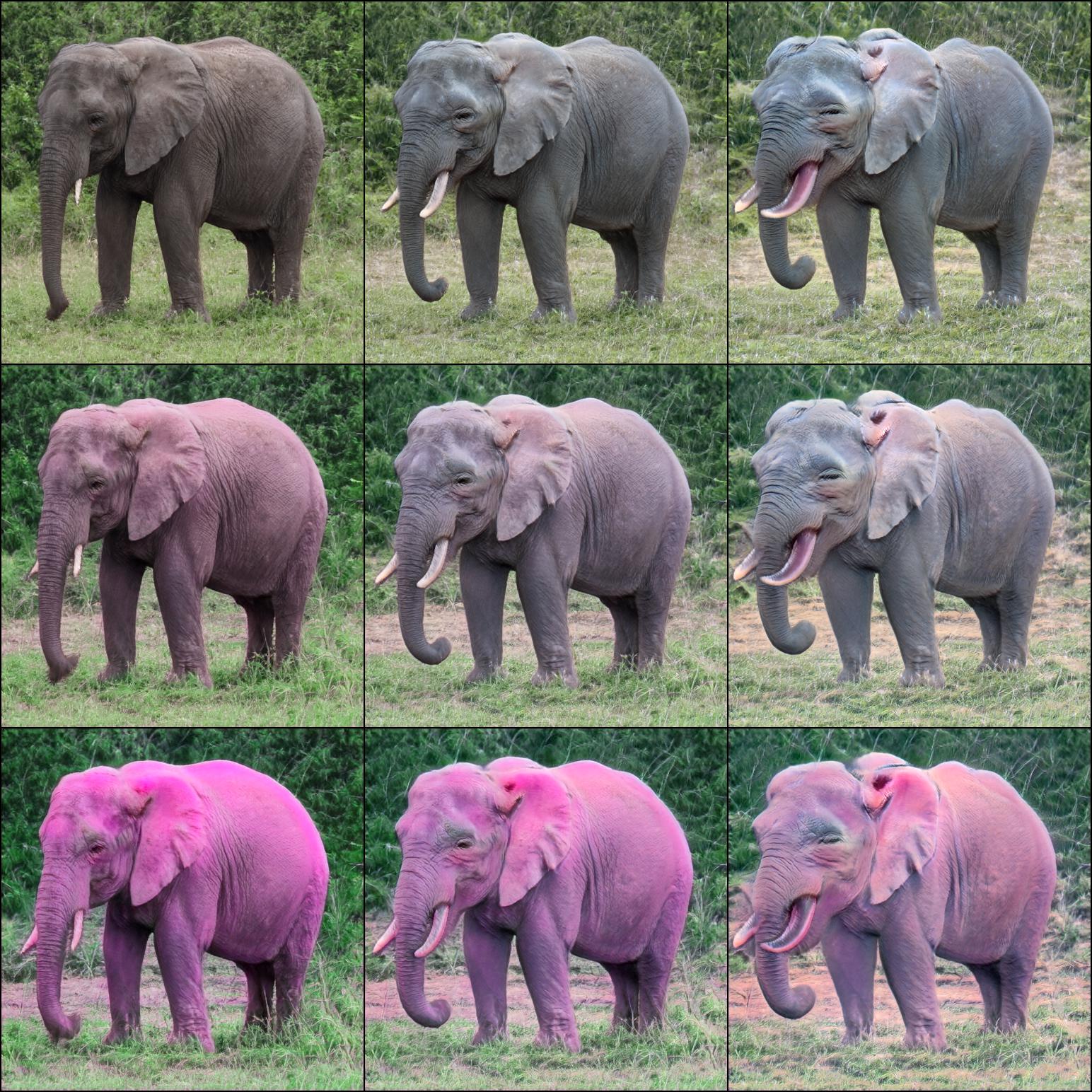}
    \end{tabular}
    \caption{Composition of factors of variation introduced to generators pretrained on LSUN Church \cite{yu2015lsun} and SD-Elephant \cite{mokady2022selfdistilled}. Following the format of \cref{fig:composition_showcase}}.
    \label{fig:church_composition_supp}
\end{figure*}

In \cref{fig:ffhq_composition_supp,fig:church_composition_supp} we provide additional qualitative results displaying compositionality in expanded generators.

\section{Additional Details}
\label{sec:supp_details}

\subsection{Training Time and Iterations}
When expanding the generator with a single new domain, our training requires roughly twice the number of iterations to obtain comparable effects. The difference is a direct result of our additional regularization terms. 
With additional domains, we observe a roughly linear relationship between the number of domains and the required training iterations. For example, the FFHQ model expanded with $105$ iterations was trained for $40K$ iterations, while the model with $400$ iterations was trained for $150K$ iterations.

Note that different training objective might require a different number of iterations. StyleGAN-NADA \cite{gal2021stylegan} specifically heavily relies on early-stopping. An ideal domain expansion method could consider this issue, and sample training objectives to apply non-uniformly. In practice, we did not observe this to be an issue, probably due to our method optimizing numerous objectives simultaneously. 

\subsection{Transformation of Loss Function}
As explained in \cref{subsec:adjusting}, transforming a given domain adaptation task to perform domain expansion requires limiting the samples latent codes. The loss function itself, in principal, is left unchanged.
This is exactly the case for MyStyle \cite{nitzan2022mystyle}. For StyleGAN-NADA \cite{gal2021stylegan}, however, we made a subtle modification to the loss function.

StyleGAN-NADA computes its loss with respect to a frozen copy of the source generator (See \cref{eq:directional_loss}). This is done in order to maintain access to the source domain, despite it vanishing from the adapted generator during training. Conversely, using our method, the source domain is preserved along the base subspace. We take advantage of this fact and modify the loss only slightly. Instead of using a frozen generator to generate images from the source domain, we simply use our expanded generator and latent codes from the base subspace.

\end{document}